%% file: main-ACL.tex
\pdfoutput=1

\documentclass[11pt]{article}

\usepackage{EMNLP2023}

\usepackage{times}
\usepackage{latexsym}

\usepackage[T1]{fontenc}

\usepackage[utf8]{inputenc}

\usepackage{microtype}

\usepackage{inconsolata}

\input{math_commands.tex}

\usepackage[utf8]{inputenc} 
\usepackage[T1]{fontenc}    

\usepackage{hyperref}       
\usepackage{url}            

\usepackage{physics}        
\usepackage{nicefrac}       
\usepackage{microtype}      
\usepackage{amsfonts}       
\usepackage{amsthm}
\usepackage{amsmath}
\usepackage{breqn}
\usepackage{braket}

\usepackage{subfig}         
\usepackage{caption}        

\usepackage{xcolor}

\usepackage{booktabs}       
\usepackage{xr}             
\usepackage{enumitem}       
\usepackage{multirow}       
\usepackage{bbm}

\usepackage{graphicx}       
\usepackage{graphics}       
\usepackage{verbatim}
\usepackage[export]{adjustbox}
\usepackage{array}
\usepackage{amssymb}
\usepackage{pifont}
\usepackage{wrapfig}

\newcommand{\method}{\textsc{NeuroSIM}}
\newcommand{\clevrmanip}{\textsc{CIM-NLI}}
\newcommand{\clevrmaniplarge}{\textsc{CIM-NLI-LARGE}}

\newcommand{\cmark}{\ding{51}}
\newcommand{\xmark}{\ding{55}}

\usepackage{algorithm2e}

\RestyleAlgo{ruled}
\usepackage{wrapfig}
\usepackage[normalem]{ulem}
\usepackage{float}
\usepackage{enumitem}
\date{}
    \makeatletter
\def\@fnsymbol#1{\ensuremath{\ifcase#1\or \dagger\or *\or
   \mathsection\or \mathparagraph\or \|\or **\or \dagger\dagger
   \or \ddagger\ddagger \else\@ctrerr\fi}}
    \makeatother

\title{Image Manipulation via Multi-Hop Instructions - A New Dataset and Weakly-Supervised Neuro-Symbolic Approach}

\author{%
    Harman Singh$^1$\quad Poorva Garg$^{1*}$
    \quad Mohit Gupta$^{1*}$ \\ \quad \textbf{Kevin Shah}$^1$\thanks{\hspace{0.3em} Equal Contribution, $^*$Equal Contribution. \\ First four authors did this work while at IIT Delhi. \\ Correspondence: \texttt{harmansingh.iitd@gmail.com, \\ parags@iitd.ac.in}} \quad \textbf{Ashish Goswami}$^{1\dagger}$ \quad \textbf{Satyam Modi}$^{1\dagger}$ \\
    \quad \textbf{Arnab Kumar Mondal}$^1$ \quad \textbf{Dinesh Khandelwal}$^2$
    \quad \textbf{Dinesh Garg}$^{2}$ \quad \textbf{Parag Singla}$^{1}$ \\
    $^1$Indian Institute of Technology Delhi \quad $^2$IBM Research AI \\
}

\begin{document}
\maketitle
\input{./sections/abstract}
\input{./sections/intro}

\input{./sections/related}
\input{./sections/model}
\input{./sections/expt}
\input{./sections/conclusion}
\input{./sections/ethics}
\input{./sections/limitations}
\input{./sections/ack}

\bibliography{references_redone}
\bibliographystyle{acl_natbib}

\newpage
\clearpage  
\appendix
\input{./sections/appendix}

\end{document}

%% file: math_commands.tex
\usepackage{amsmath,amsfonts,bm}

\def\eqref#1{equation~\ref{#1}}

\def\1{\bm{1}}

\DeclareMathAlphabet{\mathsfit}{\encodingdefault}{\sfdefault}{m}{sl}
\SetMathAlphabet{\mathsfit}{bold}{\encodingdefault}{\sfdefault}{bx}{n}



%% file: sections/abstract.tex
\begin{abstract}
We are interested in {\em image manipulation via natural language text} -- a task that is useful for multiple AI applications but requires complex reasoning over multi-modal spaces. We extend recently proposed Neuro Symbolic Concept Learning (NSCL)~\cite{NSCL}, which has been quite effective for the task of Visual Question Answering (VQA), for the task of image manipulation. Our system referred to as \method{} can perform {\em complex multi-hop reasoning} over {\em multi-object scenes} and only requires {\em weak supervision} in the form of annotated data for VQA. \method{} parses an instruction into a symbolic program, based on a Domain Specific Language (DSL) comprising of object attributes and manipulation operations, that guides its execution. We create a new dataset for the task, and extensive experiments demonstrate that \method{} is highly competitive with or beats SOTA baselines that make use of supervised data for manipulation.
\end{abstract}

%% file: sections/intro.tex
\section{Introduction}
\label{intro}
\begin{figure*}[h!]
    \centering
    \includegraphics[width=0.9\textwidth]{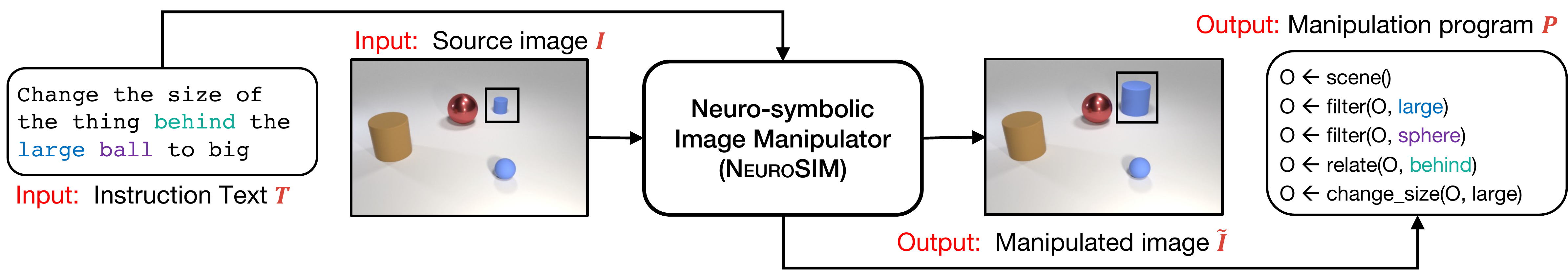}
    \caption{The problem setup. See Section \ref{intro} for more details.}
    \label{Setup}
\end{figure*}
The last decade has seen significant growth in the application of {\em neural models} to a variety of tasks including those in computer vision~\citep{chen2017deeplab,imagenet_classification}, NLP~\citep{Google_NMT}, robotics and speech~\citep{asr}. 
It has been observed that these models often lack interpretability~\citep{fan2021interpretability}, and may not always be well suited to handle complex reasoning tasks~\citep{dai2019bridging}.
On the other hand, {\em classical AI systems}
can seamlessly perform complex reasoning in an interpretable manner due to their {\em symbolic representation}~\citep{pham2007building,cai2012configuration}. But these models often lack in their ability to handle low-level representations and be robust to noise. 
{\em Neuro-Symbolic models}~\citep{NeuralLogicMachines,NSCL,VCML} overcome these limitations by combining the power of (purely) neural with (purely) symbolic representations. 
Studies~\citep{Andreas2016, hu2017learning, Johnson2017, NSCL} have shown that neuro-symbolic models have several desirable properties such as {\em modularity, interpretability,} and {\em improved generalizability}. 
 
Our aim in this work is to build neuro-symbolic models for the task of {\em weakly supervised manipulation of images comprising multiple objects, via complex multi-hop natural language instructions.} 
Specifically, we are interested in weak supervision that only uses the data annotated for VQA tasks, avoiding the high cost of getting supervised annotations in the form of target manipulated images. Our key intuition here is that this task can be solved simply by querying the manipulated representation without ever explicitly looking at the target image.
The prior work includes weakly supervised approaches~\citep{nam2018text,li2020manigan} that require textual descriptions of images during training and are limited to very simple scenes (or instructions). (See Section~\ref{sec:related} for a survey). 

Our solution builds on Neuro-Symbolic Concept Learner (NSCL) proposed by \citep{NSCL} for solving VQA. We extend this work to incorporate the notion of manipulation operations such as {\em change}, {\em add,} and {\em remove} objects in a given image. 
As one of our main contributions, we design novel neural modules and a training strategy that just uses VQA annotations as weakly supervised data for the task of image manipulation. 
The neural modules are trained with the help of {\em novel loss functions} that measure the faithfulness of the manipulated scene and object representations by accessing a separate set of {\em query networks}, interchangeably referred to as {\em quantization networks}, 
trained just using VQA data. The manipulation takes place through interpretable programs created using primitive neural and symbolic operations from a Domain Specific Language (DSL). Separately, a network is trained to render the image from a scene graph representation using a combination of $L_1$ and adversarial losses as done by~\cite{Sg2Im}. The entire pipeline is trained without any intermediate supervision. We refer to our system as Neuro-Symbolic Image Manipulator (\method).
Figure~\ref{Setup} shows an example of I/O pair for our approach. Contributions of our work are as follows:
\begin{enumerate}[leftmargin=*,noitemsep]
    \item We create \method{}, the first neuro-symbolic, weakly supervised, and interpretable model for the task of text-guided image manipulation, that does not require output images for training.
    \item We extend CLEVR~\citep{CLEVRDatasetPaper}, a benchmark dataset for VQA, to incorporate manipulation instructions and create a new dataset called as {\em Complex Image Manipulation via Natural Language Instructions} {(\clevrmanip{})}. We also create \clevrmaniplarge{} dataset to test zero-shot generalization.
    \item We provide extensive quantitative experiments on newly created \clevrmanip{}, \clevrmaniplarge{} datasets along with qualitative experiments on Minecraft \citep{NSVQA}. Despite being weakly supervised,~\method{} is highly competitive to supervised SOTA approaches including a recently proposed diffusion based model~\cite{brooks2022instructpix2pix}.
    \method{} also performs well on instructions requiring multi-hop reasoning, all while being interpretable. We publicly release our code and data~\footnote{\url{https://github.com/dair-iitd/NeuroSIM}}.
\end{enumerate}

%% file: sections/related.tex
\section{Related Work}\label{sec:related}
Table \ref{table:tech_compare} categorizes the related work across three broad dimensions - {\em problem setting}, {\em task complexity}, and {\em approach}. The problem setting comprises two sub-dimensions:
i) supervision type - {\em self, direct,} or {\em weak}, ii) instruction format - {\em text or UI-based}. The task complexity comprises of following sub-dimensions: 
ii) scene complexity -- {\em single} or {\em multiple objects}, ii) instruction complexity - {\em zero or multi-hop instructions}, iii) kinds of manipulations allowed - {\em add, remove,} or {\em change}. Finally, the approach consists of the following sub-dimensions: i) model -- {\em neural} or {\em neuro-symbolic} and ii) whether a symbolic program is generated on the way or not. 
\begin{table*}[ht]
	\centering
	\setlength\tabcolsep{1.9pt} 
	\renewcommand{\arraystretch}{1}
	\setlength\extrarowheight{2pt}
	\small{
		\begin{tabular}{llcccclccc} \hline
			\toprule
			\multirow{2}{*}{{\parbox{2cm}{{Prior Work}}}} & \multicolumn{2}{c}{Problem Setting} && \multicolumn{3}{c}{Task Complexity} && \multicolumn{2}{c}{Approach}
			\\ 
			\cline{2-3}\cline{5-7}\cline{9-10}
			& {Supervision Type} & {Instruction Format}  & & {SC} & {IC} & {Operations} && {Model} &  {Program} \\
			\toprule
			{SIMSG} & Self Supervision & UI  && MO & N/A & change, remove, add && N & \xmark \\
			{PGIM} & Direct Supervision & N/A && MO\textsuperscript{*} & N/A &  change (image level) && NS & \cmark \\
			{GeNeVA} & Direct Supervision & Text  && MO & Multi-Hop &  add && N & \xmark \\
			{TIM-GAN} & Direct Supervision & Text  && MO & Zero-Hop &  change, remove, add && N & \xmark \\
			{Dong et. al} & Weak Supervision & Text  && SO & Zero-Hop & change && N & \xmark \\
			{TAGAN} & Weak Supervision & Text  && SO & Zero-Hop &  change && N & \xmark \\
			{ManiGAN} & Weak Supervision & Text  && SO & Zero-Hop &  change && N & \xmark \\
                {InstructPix2Pix} & Pre-training + Supervision & Text  && MO & Multi-Hop &  change, remove, add && N & \xmark \\
			\method{} (ours) & Weak Supervision & Text  && MO & Multi-Hop &  change, remove, add && NS & \cmark \\
			\bottomrule
		\end{tabular}
	}
	\caption{{Comparison of Prior Work. Abbreviations (column titles) {SC}:= Scene Complexity, {IC}:=Instruction Complexity. Abbreviations (column values) MO:= Multiple Objects, MO$^*$:= Multiple Objects with Regular Patterns, SO:= Single Object, N:= Neural, NS:= Neuro-Symbolic, N/A:= Not applicable, \cmark := Yes, \xmark := No. See Section \ref{sec:related} for more details.}}
	\label{table:tech_compare}
\end{table*}
\citet{dong2017semantic}, TAGAN~\citep{nam2018text}, and ManiGAN~\citep{li2020manigan} are close to us in terms of the problem setting. These manipulate the source image using a GAN-based encoder-decoder architecture. Their weak supervision differs from ours -- We need VQA annotation, they need captions or textual descriptions. The complexity of their natural language instructions is restricted to 0-hop. Most of their experimentation is limited to single (salient) object scenes.

In terms of task complexity, the closest to us are approaches such as TIM-GAN~\citep{TIMGAN}, GeNeVA~\citep{GeNeVa}, which build an encoder-decoder architecture and work with a latent representation of the image as well as the manipulation instruction. They require a large number of manipulated images as explicit annotations for training. 

In terms of technique, the closest to our work are neuro-symbolic approaches for VQA such as ~NSVQA~\citep{NSVQA}, NSCL~\citep{NSCL}, Neural Module Networks~\citep{Andreas2016} and its extensions~\citep{hu2017learning, Johnson2017}. Clearly, while the modeling approach is similar and consists of constructing latent programs, the desired tasks are different in the two cases. Our work extends the NSCL approach  for the task of automated image manipulation. 

~\citet{jiang2021language},\citet{shi2021learning} deal with editing global features, such as brightness, contrast, etc., instead of object-level manipulations like in our case. Recent models such as InstructPix2Pix~\citep{brooks2022instructpix2pix}, DALL-E~\citep{dalle} and Imagen~\cite{imagen} on text-to-image generation using diffusion models are capable of editing images but require captions for input images; preliminary studies \citep{marcusdalle} highlight their shortcomings in compositional reasoning and handling relations. 

%% file: sections/model.tex
\section{Neuro-Symbolic Image Manipulator}
\label{section:modeldetails}
\subsection{Motivation and Architecture Overview}
\label{subsec:modelmotivation}
\begin{figure*}[t]
    \centering
    \includegraphics[width=0.7\textwidth]{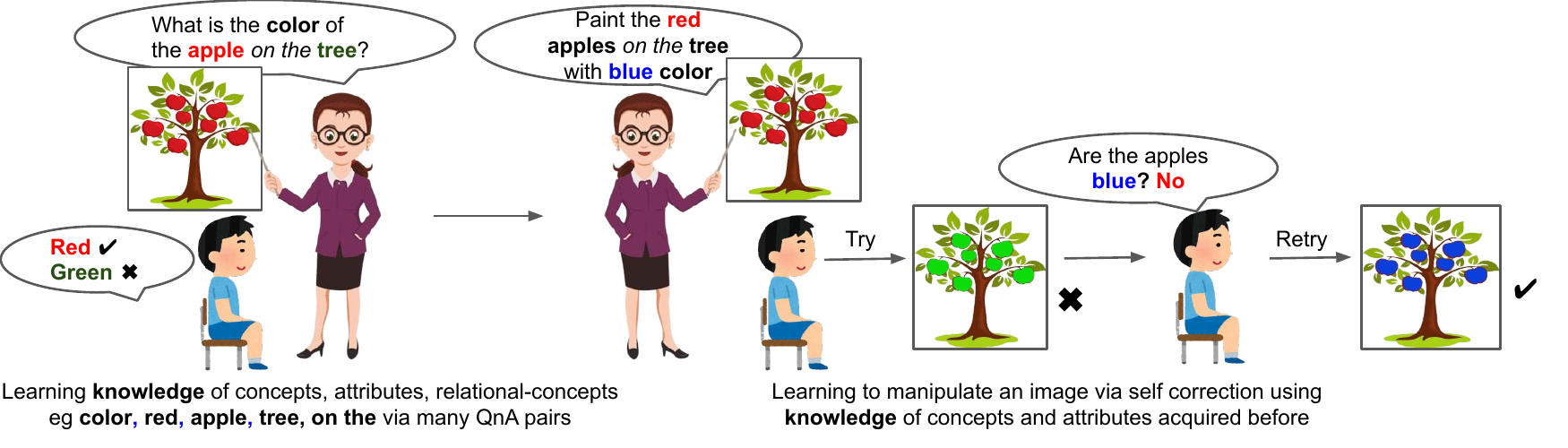}
    \vspace{-0.5mm}
    \caption{Motivating example \method{}. Best viewed under magnification. See Section \ref{subsec:modelmotivation} for more details}
    \label{fig:motivation}
\end{figure*}
The key motivation behind our approach comes from the following hypothesis: consider a learner $L$ (e.g., a neural network or the student in Fig \ref{fig:motivation}) with sufficient capacity trying to achieve the task of manipulation over Images $I$. Further, let each image be represented in terms of its properties, or properties of its constituents (e.g. objects like apple, leaf, tree, etc.  in Fig \ref{fig:motivation}), where each property comes from a finite set $S$ e.g, attributes of objects in an image. Let the learner be provided with the prior knowledge (for e.g. through Question Answering as in Fig \ref{fig:motivation}) about properties (e.g., color) and their possible values (e.g., red). Then, in order to learn the task of manipulation, it suffices to provide the learner with a {\em query network}, which given a manipulated image $\tilde{I}$ constructed by the learner via command $C$, can correctly answer questions (i.e. query) about the desired state of various properties of the constituents of the image $\tilde{I}$. The query network can be internal to the learner (e.g., the student in Fig \ref{fig:motivation} can query himself for checking the color of apples in the manipulated image). The learner can query repeatedly until it learns to perform the manipulation task correctly. Note, the learner does not have access to the supervised data corresponding to triplets of the form $(I_s,C,I_f)$, where $I_s$ is the starting image, $C$ is the manipulation command, and $I_f$ is the target manipulated image. 
Inspired by this, we set out to test this hypothesis by building a model capable of manipulating images, without target images as supervision.

\begin{figure*}[ht]
    \vspace{-0.4cm}
    \centering
    \includegraphics[keepaspectratio,width=0.9\textwidth]{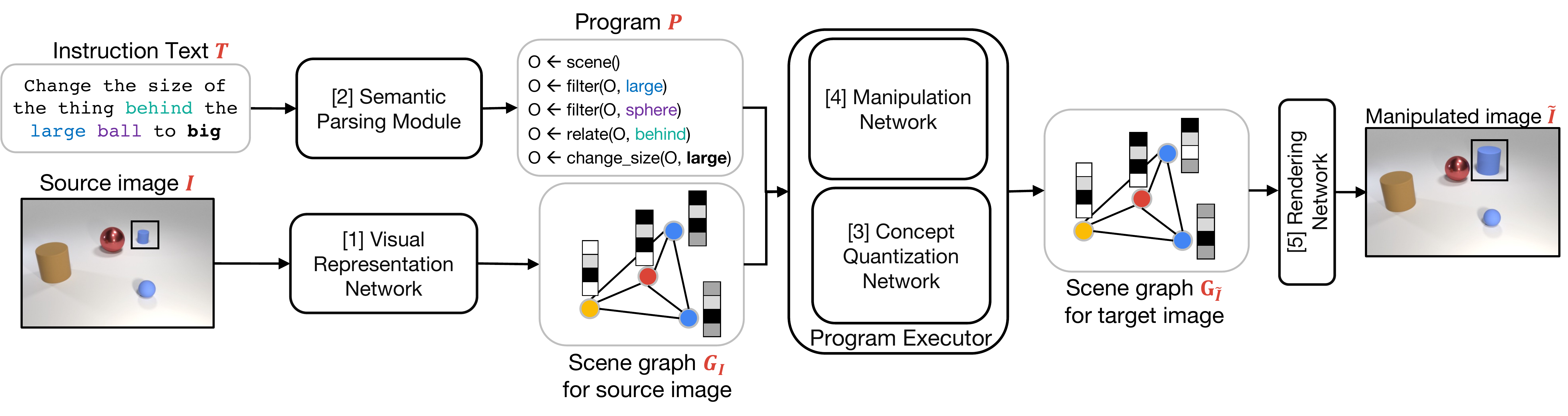}
    \vspace{-0.2cm}
    \caption{High level architecture of {\method{}}. See Section \ref{section:modeldetails} for more details.}
    \label{Pipeline}
    \vspace{-0.62cm}
\end{figure*}

Figure \ref{Pipeline} captures a high-level architecture of the proposed \method{} pipeline. \method{} allows manipulating images containing multiple objects, via complex natural language instructions.  Similar to~\citet{NSCL}, \method{} assumes the availability of a \emph{domain-specific language} (DSL) for parsing the instruction text ${T}$ into an executable program ${P}$. \method{} is capable of handling {\em addition, removal,} and {\em change} operations over image objects. 
It reasons over the image for locating where the manipulation needs to take place followed by carrying out the manipulation operation. The first three modules, namely {\em i) visual representation network, ii) semantic parser,} and {\em iii) concept quantization network} are suitably customized from the NSCL and trained as required for our purpose.  
In what follows, we describe the design and training mechanism of \method{}.
\subsection{Modules Inherited from NSCL}
\label{subsec:vqa}
{\bf 1] Visual Representation Network:} 
Given input image $I$, this network converts it into a scene graph ${G_{I}} = ({N}, {E})$. The nodes ${N}$ of this scene graph are object embeddings and the edges $E$ are embeddings capturing the relationship between pair of objects (nodes). Node embeddings are obtained by passing the bounding box of each object (along with the full image) through a ResNet-34~\citep{resnet}.  Edge embeddings are obtained by concatenating the corresponding object embeddings.
\\{\bf 2] Semantic Parsing Module:} 
The input to this module is a manipulation instruction text $T$ in natural language. Output is a {\em symbolic program} $P$ generated by parsing the input text. The symbolic programs are made of operators, that are part of our DSL (Specified in Appendix Section \ref{app:dsl}).
\\{\bf 3] Concept Quantization Network:}
Any object in an image is defined by the set of {\em visual attributes} $(A)$, and set of symbolic values $(S_a)$ for each attribute $a\in A$. E.g., attributes can be {\em shape, size,} etc. 
Different symbolic values allowed for an attribute are also known as {\em concepts}. E.g., $S_{\text{color}}= \{\text{red, blue, green,} \ldots\}$. Each visual attribute $a \in A$ is implemented via a separate neural network $f_a(\cdot)$ which takes the object embedding as input and outputs the attribute value for the object in a {\em continuous (not symbolic)} space. Let $f_{\text{color}} : \mathbb{R}^{d_{\text{obj}}} \xrightarrow{} \mathbb{R}^{d_{\text{attr}}}$ represent a neural network for the {\em color} attribute and consider $o \in \mathbb{R}^{d_{\text{obj}}}$ as the object embedding. Then, $v_{\text{color}} = f_{\text{color}}(o) \in \mathbb{R}^{d_{\text{attr}}}$ is the embedding for the object $o$ pertaining to the color attribute. Each symbolic concept $s \in S_a$ for a particular attribute $a$ (e.g., different colors) is also assigned a respective embedding in the same continuous space $\mathbb{R}^{d_{\text{attr}}}$. Such an embedding is denoted by $c_s$. These concept embeddings are initialized at random, and later on, fine-tuned during training. An attribute embedding (e.g. $v_{\text{color}}$) can be compared with the embeddings of all the concepts (e.g., $c_{\text{red}}, c_{\text{blue}}$, etc.) using cosine similarity, for the purpose of concept quantization of objects.
\\{\bf Training for VQA:} As a first step, we train the above three modules 
via a curriculum learning process \citep{NSCL}. 
The semantic parser is trained jointly with the concept quantization networks for generating programs for the question texts coming from the VQA dataset. The corresponding output programs are composed of primitive operations coming from the DSL (e.g. \textit{filter}, \textit{count}, etc.) and do not include constructs related to manipulation operations. This trains the first three modules with high accuracy on the VQA task. 
\subsection{Novel Modules and Training \method{}}
\label{subsec:neurosim_models}
\method{} training starts with three sub-modules trained on the VQA task as described in Section~\ref{subsec:vqa}. Next, we extend the original DSL to include three additional functional sub-modules within the semantic parsing module, namely \textit{add}, \textit{remove}, and \textit{change}. Refer to appendix section \ref{app:dsl} for details on the DSL. We now reset the semantic parsing module and train it again from scratch for generating programs corresponding to {\em image manipulation instruction text} $T$. Such a program is subsequently used by the downstream pipeline to reason over the scene graph ${G}_I$ and manipulate the image. In this step, the semantic parser is trained using an off-policy program search based {\texttt{REINFORCE}} \citep{williams1992simple} algorithm. Unlike the training of semantic parser for the VQA task, in this step, we \textit{do not} have any final \textit{answer like} reward supervision for training. Hence, we resort to a weaker form of supervision. In particular, consider an input instruction text $T$ and set of all possible manipulation program templates $\mathbb{P}_t$ from which one can create any actual program ${P}$ that is executable over the scene graph of the input image. For a program $P \in \mathbb{P}_t$, our reward is positive if this program $P$ selects any object (or part of the scene graph) to be sent to the manipulation networks (change/add/remove). 
See Appendix \ref{app:model_details} for more details. 
Once the semantic parser is retrained, we clamp the first three modules and continue using them for the purpose of parsing instructions and converting images into their scene graph representations. Scene graphs are manipulated using our novel module called {\em manipulation network} which is described next.
\\{\bf 4] Manipulation Network:} 
This is our key module responsible for carrying out the manipulation operations. We allow three kinds of manipulation operations -- {\em add, remove,} and {\em change}. Each of these operations is a composition of a quasi-symbolic and symbolic operation. A symbolic operation corresponds to a function that performs the required structural changes (i.e. addition/deletion of a node or an edge) in the scene graph ${G}_I$ against a given instruction. A quasi-symbolic operation is a dedicated neural network that takes the relevant part of ${G}_I$ as input and computes new representations of nodes and edges that are compatible with the changes described in the parsed instruction.
\\\textbf{(a) Change Network:} For each visual attribute $a \in A$ (e.g. \textit{shape, size, \ldots}), we have a separate {\em change neural network} that takes the pair of {\em (object embedding, embedding of the changed concept)} as input and outputs the embedding of the \textit{changed} object. This is the quasi-symbolic part of the change function, while the symbolic part is identity mapping. For e.g., let $g_{\text{color}} : \mathbb{R}^{d_{\text{obj}} + d_{\text{attr}}} \xrightarrow{} \mathbb{R}^{d_{\text{obj}}}$ represent the neural network that changes the {\em color} of an object. Consider $o \in \mathbb{R}^{d_{\text{obj}}}$ as the object embedding and $c_{\text{red}} \in \mathbb{R}^{d_{\text{attr}}}$ as the concept embedding for the {\em red} color, then $\widetilde{o} = g_{\text{color}}(o; c_{\text{red}}) \in \mathbb{R}^{d_{\text{obj}}}$ represents the changed object embedding, whose color would be {\em red}. After applying the change neural network, we obtain the changed representation of the object $\widetilde{o} = g_{a}(o; c_{s_a^*})$, where $s_a^*$ is the desired changed value for the attribute $a$. This network is trained using the following losses.
\vspace{-0.10cm}
\begin{align}
    \ell_a &=  -\sum_{\forall s \in S_a}{\mathbb{I}}_{s=s_a^*}\;\log\left[p(h_{a}\left(\widetilde{o}\right)=s)\right] \label{eq:Lc}\\[2pt]
    \ell_{\overline{a}} &= -\sum_{\forall a{'} \in A, a{'} \neq a}\sum_{\forall s \in S_{a{'}}}
                    \begin{aligned}[t]
                        &p(h_{a{'}}(o)=s)* \\
                            &\log[p(h_{a{'}}(\widetilde{o})=s)]
                        \end{aligned} \label{eq:Lnc}
\end{align}
where, $h_a(x)$ gives the concept value of the attribute $a$ (in symbolic form $s \in S_a$) for the object $x$.
The quantity $p\left(h_{a}(x)=s\right)$ denotes the probability that the concept value of the attribute $a$ for the object $x$ is equal to $s$ and is given as follows $p\left(h_{a}(x)=s\right) = {\exp^{dist(f_{a}(x),c_{s})}}/{\sum_{\widetilde{s} \in S_a}\exp^{dist(f_{a}(x),c_{\widetilde{s}})}}$ where, 
$dist(a, b) = (a^{\top}b - t_2) / t_1$ is the shifted and scaled cosine similarity, $t_1, t_2$ being constants. The first loss term $\ell_a$ penalizes the model if the (symbolic) value of the attribute $a$ for the manipulated object is different from the desired value $s_a^*$ in terms of probabilities. The second term $\ell_{\overline{a}}$, on the other hand, penalizes the model if the values of any of the other attributes $a{'}$, deviate from their original values. Apart from these losses, we also include following additional losses.
\begin{align}
\ell_{\text{cycle}} &=  \lVert o - g_{a}(\widetilde{o}; c_{\text{old}}) \rVert_{2}; \\
\ell_{\text{consistency}} & = \lVert o - g_{a}(o; c_{\text{old}}) \rVert_{2}
\end{align}
\begin{align}
\ell_{\text{objGAN}} &={ 
                \begin{aligned}[t]
                    &-\sum\nolimits_{o{'}\in O}[\log D((o{'}) \\
                    &+ \log(1-D\left(g_{a}(o{'}; c)\right))]
                \end{aligned}
                }
                \label{eq:Lobjgan}
\end{align}
where $c_{old}$ is the original value of the attribute $a$ of object $o$, before undergoing change. Intuitively the first loss term $\ell_{\text{cycle}}$ says that, changing an object and then changing it back should result in the same object. The second loss term $\ell_{\text{consistency}}$ intuitively means that changing an object $o$ that has value $c_{old}$ for attribute $a$, into a new object with the same value $c_{old}$, should not result in any change. These additional losses prevent the change network from changing attributes which are not explicitly taken care of in earlier losses (\ref{eq:Lc}) and (\ref{eq:Lnc}). For e.g., rotation or location attributes of the objects that are not part of our DSL. We also impose an adversarial loss $\ell_{\text{objGAN}}$ to ensure that the new object embedding $\widetilde{o}$ is from the same distribution as real object embeddings. See Appendix \ref{app:model_details} for more details.
\\\textbf{(b) Remove Network:} This network takes the scene graph ${G}_I$ of the input image
and removes the subgraph from $G_{I}$ that contains the nodes (and incident edges) corresponding to the object(s) that need to be removed,
 and returns a new scene graph ${G}_{\widetilde{I}}$ which is reduced in size. The quasi-symbolic function for the remove network is identity.
\\\textbf{(c) Add Network:} For adding a new object into the scene, {\em add network} requires the symbolic values of different attributes, say $\{{s}_{a_1}, {s}_{a_2}, \ldots, s_{a_{k}}\}$, for the new object, e.g., $\{ \text{red, cylinder,} \ldots\}$. It also requires
the spatial relation $r$ (e.g. $\text{RightOf}$) of the new object with respect to an existing object in the scene. The add function first predicts the object (node) embedding $\widetilde{o}_{\text{new}}$ for the object to be added, followed by predicting edge embeddings for new edges incident on the new node. New object embedding is obtained as follows: $\widetilde{o}_{\text{new}} = g_{\text{addObj}}(\{c_{s_{a_1}}, c_{s_{a_2}}, \cdots, c_{s_{a_k}} \}, o_{rel}, c_r)$ where,
$o_{rel}$ is the object embedding of an existing object, relative to which the new object's position $r$ is specified. For each existing objects $o_i$ in the scene, an edge $\widetilde{e}_{\text{new}, i}$ is predicted between the newly added object $\widetilde{o}_{\text{new}}$ and existing object $o_i$ in following manner: $\widetilde{e}_{\text{new}, i}= g_{\text{addEdge}}(\widetilde{o}_{\text{new}}, o_i)$. Functions $g_{\text{addObj}}(\cdot)$ and $g_{\text{addEdge}}(\cdot)$ are quasi-symbolic operations. Symbolic operations in {\em add network} comprise adding the above node and the incident edges into the scene graph. 

The {\em add network} is trained in a {\em self-supervised} manner. For this, we pick a training image and create its scene graph. Next, we randomly select an object $o$ from this image and quantize its concepts, along with a relation with any other object $o_i$ in the same image. We then use our {\em remove} network to remove this object $o$ from the scene. Finally, we use the quantized concepts and the relation that were gathered above and add this object $o$ back into the scene graph using $g_{\text{addObj}}(\cdot)$ and $g_{\text{addEdge}}(\cdot)$. Let the embedding of the object after adding it back is $\widetilde{o}_{\text{new}}$. The training losses are as follows:
\vspace{-0.3cm}
\begin{flalign}
	\ell_{\text{concepts}} &= \hspace{-0.08cm}  -\sum\limits_{j=1}^{k} \log(p(h_{a_{j}}(\widetilde{o}_{\text{new}})=s_{a_j})) && \label{eq:Laddobjconcepts} \\
	\ell_{\text{relation}} &=  \hspace{-0.08cm} - \log(p(h_{\text{r}}(\widetilde{o}_{\text{new}}, o_i)=r)) && \label{eq:Laddrelconcepts} \\
	\ell_{\text{objSup}} &=  \lVert o - \widetilde{o}_{\text{new}} \rVert_{2}  \label{eq:LaddobjSupervised} &&\\
	\ell_{\text{edgeSup}} &=  \sum\nolimits_{ i \in O}\lVert e_{\text{old}, i} - \widetilde{e}_{\text{new},i} \rVert_{2} && \label{eq:LaddedgeSupervised} \\
	\ell_{\text{edgeGAN}} &= -\hspace{-0.1cm}\sum\limits_{\forall i \in O} [\log D(\{o;e_{\text{old},i};o_i\}) + && \nonumber \\ 	
   &\log(1-D\left(\{\widetilde{o}_{\text{new}};\widetilde{e}_{\text{new},i};o_i\}\right))] 
	\label{eq:Ledgegan}
\end{flalign}

where $s_{a_j}$ is the  required (symbolic) value of the attribute $a_j$ for the original object $o$, and $r$ is the required relational concept. $O$ is the set of the objects in the image, $e_{\text{old}, i}$ is the edge embedding for the edge between original object $o$ and its neighboring object $o_i$. Similarly, $\widetilde{e}_{\text{new},i}$ is the corresponding embedding of the same edge but after when we have (removed + added back) the original object. The loss terms $\ell_{\text{concepts}}$ and $\ell_{\text{relation}}$ ensure that the added object comprises desired values of attributes and relation, respectively. Since we had first removed and then added the object back, we already have the original edge and object representation, and hence we use them in loss terms given in \eqref{eq:LaddedgeSupervised}. We use adversarial loss \eqref{eq:Ledgegan} for generating real (object, edge, object) triples and also a loss similar to \eqref{eq:Lobjgan} for generating real objects. 
\subsection{Image Rendering from Scene Graph}\label{subsec:image_rendering_network}
{\bf 5] Rendering Network:}
Following Johnson et al.~\citeyearpar{Sg2Im}, the scene graph for an image is first generated using the {\em visual representation network}, which is the processed by a GCN 
and passed through a mask regression network followed by a box regression network to generate a coarse 2-dimensional structure (scene layout). A Cascaded Refinement Network \citep{crn} is then employed to generate an image from the scene layout. A {\em min-max adversarial training procedure} is used to generate realistic images, using a patch-based and object-based discriminator.

%% file: sections/expt.tex
\section{Experiments}
\label{sec_experiments}

\begin{figure*}[h]
	\centering
	\includegraphics[keepaspectratio,width=1.0\textwidth,trim={0 0 0 0},clip]{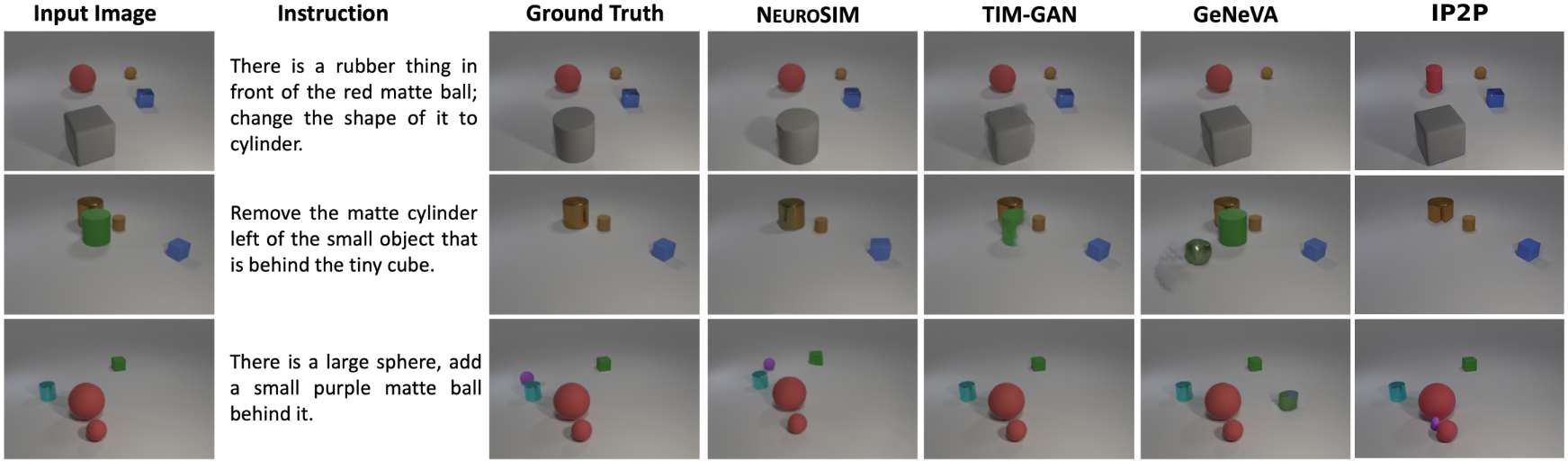}
	\caption{Visual comparison of \method{} with various baselines. See Section \ref{section:qualitative_analysis} for more details.}
	\label{fig:visual_comparisons}
        \vspace{-0.5cm}
\end{figure*}
\noindent
{\bf Datasets:}
Among the existing datasets, CSS~\citep{Vo_2019_CVPR} contains simple $0$-hop instructions and is primarily designed for the text-guided image retrieval task. 
Other datasets such as i-CLEVR \citep{GeNeVa} and CoDraw are designed for iterative image editing. i-CLEVR contains only "add" instructions and CoDraw doesn't contain multi-hop instructions. Hence we created our own {\em multi-object multi-hop instruction} based image manipulation dataset, referred to as \clevrmanip{}. This dataset was generated with the help of CLEVR toolkit~\citep{CLEVRDatasetPaper}. 
\clevrmanip{} consists of {(Source image $I$, Instruction text $T$, Target image $\widetilde{I}^*$)} triplets.  The dataset contains a total of $18K$, $5K$, $5K$ unique images and $54K$, $14K$, $14K$ instructions in the {\em train, validation and test} splits respectively. Refer to Appendix \ref{app:datasets} for more details about the dataset generation and dataset splits.

\noindent
{\bf Baselines:}
We compare our model with purely supervised approaches such as TIM-GAN~\citep{TIMGAN}, GeNeVA~\citep{GeNeVa} and InstructPix2Pix~\citep{brooks2022instructpix2pix}. In order to make a fair and meaningful comparison between the two kinds (supervised and our, weakly-supervised) approaches, we carve out the following set-up. Assume the cost required to create one single annotated example for image manipulation task be $\alpha_{m}$ while the corresponding cost for the VQA task be $\alpha_v$. Let $\alpha={\alpha_m}/{\alpha_v}$. Let $\beta_m$ be the number of annotated examples required by a supervised baseline for reaching a performance level of $\eta_m$ on the image manipulation task. Similarly, let $\beta_v$ be the number of annotated VQA examples required to train \method{} to reach the performance level of $\eta_v$. Let $\beta = {\beta_m}/{\beta_v}$. We are interested in figuring out the range of $\beta$ for which performance of our system  ($\eta_v$) is at least as good as the baseline ($\eta_m$). Correspondingly we can compute the ratio of the labeling effort required, i.e., $\alpha*\beta$,  to reach these performance levels.

If $\alpha*\beta > 1$, our system achieves the same or better performance, with lower annotation cost. Weakly supervised models~\citep{li2020manigan,nam2018text} are designed for a problem setting different from ours -- single salient object scenes, simple $0$-hop instructions (Refer Section~\ref{sec:related} for details). Further, they require paired images and their textual descriptions as annotations. We, therefore, do not compare with them in our experiments.
See Appendix \ref{computational_resources}, \ref{hyperparams} for computational resources and hyperparameters respectively.\\
\noindent
\\{\bf Evaluation Metrics:} 
For evaluation on image manipulation task, we use three metrics - i) {\em FID}, ii) Recall@$k$, and iii) Relational-similarity (rsim). FID \citep{FID} measures the realism of the generated images. We use the implementation proposed in ~\citet{parmar2021cleanfid} to compute FID. Recall@$k$ measures the semantic similarity of gold manipulated image $\widetilde{I}^*$ and system produced manipulated image $\widetilde{I}$. For computing Recall@$k$, we follow ~\citet{TIMGAN}, i.e. we use $\widetilde{I}$ as a query and retrieve images from a corpus comprising the entire test set. {\em rsim} measures how many of the ground truth relations between the objects are present in the generated image. We follow ~\cite{GeNeVa} to implement rsim metric that uses predictions from a trained object-detector (Faster-RCNN) to perform relation matching between the scene-graphs of ground-truth and generated images.
\begin{table}[ht]
    \setlength\tabcolsep{1.8pt}
    \renewcommand{\arraystretch}{0.9}
    \setlength\extrarowheight{2pt}
    \raggedright
    \small{
    \begin{tabular}{lrrrrrrrrrrr}
    \toprule[1.0pt]
     \multirow{3}{*}{{Method}} 
    && \multicolumn{4}{c}{$\beta =0.054$} 
    && \multicolumn{4}{c}{$\beta =0.54$} \\

    \cline{3-6} \cline{8-11}
    && {FID} & {$R1$} & {$R3$} & rsim && {FID}  & {$R1$} & {$R3$} & rsim\\ \midrule
    {\small GeNeVA} && $42.6$  & $6.6$ & $58.7$ & $80.1$ && $28.5$ & $4.6$ & $64.4$ & $84.9$\\
    {\small TIM-GAN} && $24.2$   & $31.9$ & $74.2$ & $88.4$ && $22.7$ & $58.1$ & $90.2$ & $94.0$\\
    {\small IP2P} && $ 3.4$  &$40.6$ & $ 77.0$ & $88.8$ && $ 2.2$ & $49.2$ & $ 84.8$ & $94.5$\\
    {\small \method{}} && $35.0$ & $45.3$ & $65.5$ & $91.3$ && $35.1$ &$45.5$ & $66.7$ & $91.5$\\ 
    \bottomrule[1.0pt]
    \end{tabular}
    }
    \caption{Performance comparison of \method{} with TIM-GAN and GeNeVA, and InstructPix2Pix (IP2P) with 10\% data ($\beta =0.054$) and full data ($\beta =0.54$). 
    We always use $100K$ VQA examples (5K Images, 20 questions per image) for our weakly supervised training.$R1$, $R3$ correspond to Recall@1,3 respectively. {FID}: lower is better; {Recall}/{rsim}: higher is better. See Section \ref{perf_varying_size} for more details.}
   \label{quantitative_comparisons_combined}
    \vspace{-0.5cm}
\end{table}

\subsection{Performance with varying Dataset Size}
\label{perf_varying_size}
Table \ref{quantitative_comparisons_combined} compares the performance of \method{} other SoTA methods two level of $\beta$ $0.054$ and $0.54$ representing use of 10\% and 100\% samples from \clevrmanip{}.  
Despite being weakly supervised, \method{} performs significantly better than the baselines with just 10k data samples (especially TIM-GAN) and not too far from diffusion model based IP2P in full data setting, using the $R@1$ performance metric. This clearly demonstrates the strength of our approach in learning to manipulate while only making use of VQA annotations. We hypothesize that, in most cases, \method{} will be preferable since we expect the cost of annotating an output image for manipulation to be significantly higher than the cost of annotating a VQA example.
To reach the performance of the \method{} in a low data regime, TIM-GAN requires a larger number of expensive annotated examples (ref. Table \ref{quantitative_comparisons_combined_supp} in Appendix). The FID metric shows similar trend across dataset sizes and across models. The FID scores for \method{} could potentially be improved by jointly training VQA module along with image decoder and is a future direction.\\

We evaluate InstructPix2Pix (IP2P)~\cite{brooks2022instructpix2pix}, a state-of-the-art pre-trained diffusion model for image editing, in a zero-shot manner on the CIM-NLI dataset. Considering its extensive pre-training, we expect IP2P to have learned the concepts present in the CIM-NLI dataset. 
In this setting IP2P achieves a FID score of $33.07$ and $R@1$ score of $7.48$ illustrating the limitations of large-scale models in effectively executing complex instruction-based editing tasks without full dataset fine-tuning. Table \ref{quantitative_comparisons_combined} contains the results obtained by IP2P after fine-tuning for 16k iterations on CIM-NLI dataset.

\begin{table}[ht]
\centering
\setlength\tabcolsep{2pt}
\setlength\extrarowheight{2pt}
\centering
\small{
\begin{tabular}{lrrlrrrr}
    \toprule
    \multirow{2}{*}{{Method}} &\multicolumn{2}{c}{Larger Scenes} && \multicolumn{3}{c}{Hops}\\
    \cline{2-3} \cline{5-7}
    & $R1$ & $R3$ & &  \multicolumn{1}{r}{$ZH$} & \multicolumn{1}{r}{$MH$}\ & \multicolumn{1}{r}{$\triangle$}\ \\
    \midrule
    {\small GeNeVA $54K$}    & $5.0$ & $65.8$    & & $6.3$ & \multicolumn{1}{r}{$6.4$} & {\scriptsize\color{blue}(+0.1)} \\
    {\small GeNeVA $5.4K$}  & $8.2$  & $64.6$   &  & $8.5$ & \multicolumn{1}{r}{$9.9$} &{\scriptsize\color{blue}(+1.4)} \\
    {\small TIM-GAN $54K$}    & $66.3$ & $92.4$  & & $84.0$ & \multicolumn{1}{r}{$76.2$} &{\scriptsize\color{blue}(-7.8)} \\ 
    {\small TIM-GAN $5.4K$}      & $30.2$ & $80.7$ & & $56.4$ & \multicolumn{1}{r}{$41.6$} &{\scriptsize\color{blue}(-14.8)} \\ 
    {\small IP2P $54K$}    & $69.2$ & $99.8$  & & $72.5$ & \multicolumn{1}{r}{$67.4$} &{\scriptsize\color{blue}(-5.1)} \\ 
     {\small IP2P $5.4K$}      & $64.9$ & $99.4$ & & $69.3$ & \multicolumn{1}{r}{$54.8$} &{\scriptsize\color{blue}(-14.5)} \\ 
    {\small \method{} $5.4K$}   & $63.7$ & $89.1$  & & $64.5$ & \multicolumn{1}{r}{$63.0$} &{\scriptsize\color{blue}(-1.5)} \\ 
    \bottomrule[1.0pt] 
\end{tabular}
}
\caption{(Left) Performance on generalization to Larger Scenes. (Right) $R1$ results for $0$-hop (ZH) vs multi-hop (MH) instruction-guided image manipulation. See Sections \ref{expt:multi_hop} and \ref{results:combgen} for more details.}
\label{perf_with_hops_and_larger_scenes}
\vspace{-0.5cm}
\end{table}

\subsection{Performance versus Reasoning Hops}
\label{expt:multi_hop}

Table \ref{perf_with_hops_and_larger_scenes} (right) compares baselines with \method{} for performance over instructions requiring zero-hop (ZH) versus multi-hop
($1-3$ hops) (MH) reasoning. Since there are no \textit{Add} instructions with ZH, we exclude them from this experiment for the comparison to be meaningful. 
GeNeVA performs abysmally on both ZH as well as MH.
We see a significant drop in the performance of both TIM-GAN and IP2P when going from ZH to MH instructions, both for training on $5.4K$, as well as, $54K$ datapoints. In contrast, \method\ trained on $10\%$ data, sees a performance drop of only $1.5$ points showing its robustness for complex reasoning tasks.

\subsection{Zero-shot Generalization to Larger Scenes}
\label{results:combgen}
We developed another dataset called \clevrmaniplarge{}, consisting of scenes having $10-13$ objects (See Appendix \ref{app:datasets} for details). We study the combinatorial generalization ability of \method{} and the baselines when the models are trained on \clevrmanip{} containing scenes with $3-8$ objects only and evaluated on \clevrmaniplarge{}. Table \ref{perf_with_hops_and_larger_scenes} captures such a comparison. \method{} does significantly better, i.e., 33 pts (R1) than TIM-GAN and 
is competitive with IP2P when trained on $10\%$ $(5.4K$data points$)$ of \clevrmanip{}. We do see a drop in performance relative to baselines when they are trained on full (54K) data, but this is expected as effect of supervision takes over, and ours is a weakly supervised model. Nevertheless, this experiments demonstrates the effectiveness of our model for zero-shot generalization, despite being weakly sueprvised.

\subsection{Qualitative Analysis and Interpretability}
\label{section:qualitative_analysis}
Figure \ref{fig:visual_comparisons} shows anecdotal examples for visually comparing \method{} with baselines. Note, GeNeVA either performs the wrong operation on the image (row \#1, 2, 3) or simply copies the input image to output without any modifications. TIM-GAN often makes semantic errors which show its lack of reasoning (row \#3) or make partial edits (row \#1). IP2P also suffers from this where it edits incorrect object (row \#1,2). Compared to baselines, \method{} produces semantically more meaningful image manipulation. \method{} can also easily recover occluded objects (row \#4). For more results, see Appendix~\ref{app:qualitative}, \ref{app:errors}. \method{} produces interpretable output programs, showing the steps taken by the model to edit the images, which also helps in detecting errors (ref. Appendix \ref{app:interpretable_programs}).  


\subsection{Evaluating Manipulated Scene Graph}
\label{results:scene_graph_quality}
We strongly believe {\em image rendering module} of \method{} pipeline and {\em encoder} modules used for computing Recall$@k$ add some amount of inefficiencies resulting in lower $R1$ and $R3$ scores for us. Therefore, we decide to assess the quality of manipulated scene graph $G_{\widetilde{I}}$. 
\begin{wraptable}{r}{0.18\textwidth}
    \centering
    \setlength\tabcolsep{1.4pt}
    \small{
    \begin{tabular}{lrrrr}
    \toprule[1.0pt]
    {Method}   & $R1$   &  $R3$  \\ \midrule
    Text-Only  & $0.2$ & $0.4$ \\ 
    Image-Only & $34.1$ & $83.6$ \\ 
    Concat     & $39.5$ & $86.9$ \\      
    TIRG  & $34.8$ & $84.6$  \\
    \method{} & $85.8$ & $92.9$ \\                 
    \bottomrule[1.0pt] 
    \end{tabular}
    }
    \caption{$G_{\widetilde{I}}$ Quality via image retrieval.} 
    \label{quantitative_comparisons}
\end{wraptable}
For this, we consider the {\em text guided image retrieval} task proposed by \cite{Vo_2019_CVPR}. In this task, an image from the database has to be retrieved which would be the closest match to the desired manipulated image. Therefore, we use our manipulated scene graph
$G_{\widetilde{I}}$ as the latent representation of the input instruction and image for image retrieval. We retrieve images from the database based on a novel {\em graph edit distance} between \method{} generated $G_{\widetilde{I}}$ of the desired manipulated images, and scene graphs of the images in the database. This distance is defined using the {\em Hungarian algorithm}~\citep{Hungarian} with a simple cost defined between any 2 nodes of the graph (ref. Appendix \ref{supp:experiments} for details).  
Table \ref{quantitative_comparisons} captures the performance of \method{} and other popular baselines for the image retrieval task. \method{} significantly outperforms supervised learning baselines by a margin of $\sim$ $50\%$ without using output image supervision, demonstrating that \method{} meaningfully edits the scene graph. 
Refer to Section \ref{app:hum_eval} for human evaluation results and Appendix Section \ref{supp:experiments}-\ref{supp:end2end}, \ref{supp:ablations}, for more results including results on Minecraft dataset and ablations.

\subsection{A Hybrid Approach using \method{}}
\label{hybrid-model}
From Table 3, we observe that both TIM-GAN and IP2P suffer a significant drop in performance when moving from ZH to MH instructions, whereas \method{} is fairly robust to this change.
Further, we note that the manipulation instructions in our dataset are multi-hop in terms of reasoning, but once an object of interest is identified, the actual manipulation operation can be seen as single hop.
We use this observation to design a hybrid supervised baseline that utilizes the superior reasoning capability of \method{} and high quality editing and generation capabilities of IP2P.  

We take the CIM-NLI test set and parse the text-instructions through our trained semantic-parser to obtain the object embeddings over which the manipulation operation is to be performed. We utilize our trained query networks to obtain the symbolic attributes such as color, shape, size and material of the identified object. Using these attributes we simplify a complex multi-hop instruction into a simple instruction with 0 or 1 hops using a simple template based approach (see Appendix Section~\ref{simplifiy_MH} for details). These simplified instructions are fed to the fine-tuned IP2P model to generate the edited images. We refer to our hybrid approach as IP2P-NS where NS refers to Neuro-Symbolic. 
Table \ref{quantitative_comparisons_combined_hybrid_model} presents the results. 
We find that there is a clear advantage of using a hybrid neuro-symbolic model integrating \method{} with IP2P.
We see a significant gain on FID, recall, rsim when we use the hybrid approach, especially in the low resource setting ($\beta=0.054$). Compared to IP2P, the hybrid neuro-symbolic approach results in better FID, recall and rsim scores,
except a small drop in R1 for $\beta=0.54$ setting. This opens up the possibility of further exploring such hybrid models in future for improved performance (in the supervised setting).

\begin{table}[ht]
    \setlength\tabcolsep{1.8pt}
    \renewcommand{\arraystretch}{0.9}
    \setlength\extrarowheight{2pt}
    \raggedright
    \small{
    \begin{tabular}{lrrrrrrrrrrrr}
    \toprule[1.0pt]
     \multirow{3}{*}{{Method}} 
    && \multicolumn{4}{c}{$\beta =0.054$} 
    && \multicolumn{4}{c}{$\beta =0.54$} \\

    \cline{3-6} \cline{8-11}
    && {FID} & {$R1$} & {$R3$} & rsim && {FID}  & {$R1$} & {$R3$} & rsim\\ \midrule
    {\small IP2P} && $ 3.4$  &$40.6$ & $ 77.0$ & $88.8$ && $ 2.2$ & $49.2$ & $ 84.8$ & $94.5$\\
    {\small \method{}} && $35.0$ & $45.3$ & $65.5$ & $91.3$ && $35.1$ &$45.5$ & $66.7$ & $91.5$\\ 
    {\small IP2P-NS} && $1.96$  &$45.5$ & $ 83.2$ & $94.0$ && $ 1.8$ & $48.0$ & $ 85.5$ & $95.6$\\
    \bottomrule[1.0pt]
    \end{tabular}
    }
    \caption{Comparison between IP2P-NS and IP2P.}
   \label{quantitative_comparisons_combined_hybrid_model}
\end{table}

\subsection{Human Evaluation}
\label{app:hum_eval}
For the human evaluation study, we presented 10 evaluators with a set of five images each, including: The input image, the ground-truth image and manipulated images generated by \method{} 5.4K, TIM-GAN 54K, and IP2P 54K. Images generated by the candidate models were randomly shuffled to prevent any bias. Evaluators were asked two binary questions, each requiring a 'yes' (1) or 'no' (0) response, to assess the models: (Q1) Does the model perform the desired change mentioned in the input instruction?, (Q2) Does the model not introduce any undesired change elsewhere in the image? Refer to Appendix Section \ref{hum_eval_appendix_with_details} for more details about exact questions and the human evaluation process. 

The average scores from the evaluators across different questions can be found in Table~\ref{tab:human_eval_scores}. The study achieved a high average Fleiss’ kappa score~\citep{fleiss2013statistical} of $0.646$, indicating strong inter-evaluator agreement. Notably, \method{} (5.4K) outperforms TIM-GAN and IP2P (54K) in Q1 suggesting its superior ability to do reasoning, and identify the relevant object as well as affect the desired change.In contrast, TIM-GAN and IP2P score significantly better in Q2, demonstrating their ability not to introduce unwanted changes elsewhere in the image, possibly due to better generation quality compared to \method{}. 

\begin{table}
    \setlength\tabcolsep{9pt}
    \renewcommand{\arraystretch}{1.2}
    \centering
    \begin{tabular}{l@{\hspace{6pt}}c@{\hspace{7pt}}cc}
        \toprule
        {Qn.} & {\method{}} & {TIM-GAN} & {IP2P}\\
        {} & 5.4K & 54K & 54K\\
        \midrule
        \texttt{Q1}     & $0.41$  & $0.27$ & $0.25$\\
        \midrule
        \texttt{Q2}     & $0.33$  & $0.84$ & $0.78$\\
        \bottomrule
    \end{tabular}
\caption{Human evaluation comparing various models.}
\label{tab:human_eval_scores}
\vspace{-0.5cm}
\end{table}

%% file: sections/conclusion.tex
\section{Conclusion}
We present a neuro-symbolic, interpretable approach \method\ to solve image manipulation task 
using weak supervision in the form of VQA annotations.
Our approach can handle multi-object scenes with complex instructions requiring multi-hop reasoning, and solve the task without any output image supervision. We also curate a dataset of image manipulation and demonstrate the potential of our approach compared to supervised baselines. Future work includes understanding the nature of errors made by \method{}, having a human in the loop to provide feedback to the system for correction, and experimenting with real image datasets.

%% file: sections/ethics.tex
\section{Ethics Statement}

\label{sec_impact}
All the datasets used in this paper were synthetically generated and do not contain any personally identifiable information or offensive content. The ideas and techniques proposed in this paper are useful in designing interpretable natural language-guided tools for image editing, computer-aided design, and video games. One of the possible adverse impacts of AI-based image manipulation is the creation of {\em deepfakes}~\cite{vaccari2020deepfakes} (using deep learning to create fake images). To counter deepfakes, several researchers~\cite{dolhansky2020deepfake, mirsky2021creation} have also looked into the problem of detecting real vs. fake images. 

%% file: sections/limitations.tex
\section{Limitations}
A limitation of our approach is that when transferring to a new domain, having different visual concepts requires not only learning new visual concepts but also the DSL needs to be redefined. Automatic learning of DSL from data has been explored in some prior works \cite{ellis2021dreamcoder, ellisLibraries}, and improving our model using these techniques are future work for us. We can also use more powerful graph decoders for image generation, for improved image quality, which would naturally result in stronger results on image manipulation.
\label{limitations}

%% file: sections/ack.tex
\section*{Acknowledgements}
We thank anonymous reviewers for their insightful suggestions that helped in greatly improving our paper. We also thank Rushil Gupta and other members of IIT Delhi DAIR group for their helpful comments and suggestions on this work. This work was supported by an IBM AI Horizons Network (AIHN) grant. We thank IIT Delhi HPC facility\footnote{\emph{http://supercomputing.iitd.ac.in}}, IBM cloud facility, and IBM Cognitive Computing Cluster (CCC) for computational resources. Ashish Goswami is a PhD student at Yardi-ScAI@IIT-Delhi and is supported by Yardi School of AI Publication Grant. Parag Singla was supported by the DARPA Explainable Artificial Intelligence (XAI) Program with number N66001-17-2-4032, IBM AI Horizon Networks (AIHN) grant and IBM SUR awards. Any opinions, findings, conclusions or recommendations expressed in this paper are those of the authors and do
not necessarily reflect the views or official policies, either expressed or implied, of the funding agencies.

%% file: sections/appendix.tex
\hspace{-0.4cm}\textbf{\LARGE Appendix}

\section{Domain Specific Language (DSL)}
\label{app:dsl}
Table \ref{table_dsl} captures the DSL used by our \method{} pipeline. The first $5$ constructs in this table are common with the DSL used in \citet{NSCL}.
The last $3$ operations (\texttt{Change, Add,} and \texttt{Remove}) were added by us to allow for the manipulation operations.
\begin{table*}[ht]
\centering
\setlength\tabcolsep{4pt} 
\renewcommand{\arraystretch}{1.3}
\setlength\extrarowheight{2pt}
{
\begin{tabular}{lll}
\toprule
Operation & \parbox{5.5cm}{Signature [\textit{Output} $\leftarrow$ \textit{Input}])} &  \parbox{6.5cm}{Semantics}\\
\midrule
\texttt{Scene} & ObjSet $\leftarrow$ () & Returns all objects in the scene.\\
\midrule
\texttt{Filter} & \parbox{5.5cm}{ObjSet $\leftarrow$ (ObjSet, ObjConcept)} & \parbox{6.5cm}{Filter out a set of objects from ObjSet that have a concept (e.g. red) specified in ObjConcept.}\\
\midrule
\texttt{Relate} & \parbox{5.5cm}{ObjSet $\leftarrow$ (ObjSet, RelConcept, Obj)} & \parbox{6.5cm}{Filter out a set of objects from ObjSet that have concept specified relation concept (e.g. RightOf) with object Obj.}\\
\midrule
\texttt{Query} & \parbox{5.5cm}{ObjConcept $\leftarrow$ (Obj, Attribute)} & \parbox{6.5cm}{Returns the Attribute value for the object Obj.}\\
\midrule
\texttt{Exist} & \parbox{5.5cm}{Bool $\leftarrow$ (ObjSet)} & \parbox{6.5cm}{Checks if the set ObjSet is empty.}\\
\midrule
\texttt{Change} & \parbox{5.5cm}{Obj $\leftarrow$ (Obj, Concept)} & \parbox{6.5cm}{Changes the attribute value of the input object (Obj), corresponding to the input concept, to Concept}\\
\midrule
\texttt{Add} & \parbox{5.5cm}{Graph $\leftarrow$ \\ (Graph, RelConcept, Obj, ConceptSet)} & \parbox{6.5cm}{Adds an object to the input graph, generating a new graph having the object with attribute values as ConceptSet, and present in relation RelConcept of the input Obj}\\
\midrule
\texttt{Remove} & \parbox{5.5cm}{Graph $\leftarrow$ (Graph, ObjSet)} & \parbox{6.5cm}{Removes the input objects and their edges from the input graph to output a new graph}\\
\bottomrule

\end{tabular}
}
\caption{Extended Domain Specific Language (DSL) used by \method{}.}
\label{table_dsl}
\end{table*}
Table \ref{table_type_system} shows the type system used by the DSL in this work. The first $5$ types are inherited from \cite{NSCL} while the last one is an extension of the type system for handling the inputs to the \texttt{Add} operator.
\begin{table*}[ht]
\centering
\setlength\tabcolsep{4pt} 
\renewcommand{\arraystretch}{1.3}
\setlength\extrarowheight{2pt}
{
\begin{tabular}{ll}
\toprule
Type & \parbox{5.5cm}{Remarks}\\
\midrule
\texttt{ObjConcept} & Concepts for any given object, such as {\em blue, cylinder,} etc.\\
\texttt{Attribute} & Attributes for any given object, such as {\em color, shape,} etc.\\
\texttt{RelConcept} & Relational concepts for any given object pair, such as {\em RightOf, LeftOf,} etc.\\
\texttt{Object} & Depicts a single object\\
\texttt{ObjectSet} & Depicts multiple objects\\
\texttt{ConceptSet} & A set of elements of ObjConcept type\\

\bottomrule

\end{tabular}
}
\caption{Extended type system for the DSL used by \method{}.}
\label{table_type_system}
\end{table*}

\section{Dataset Details}
\label{app:datasets}
We use CLEVR dataset and CLEVR toolkit (code to generate the dataset). These are public and are under CC and BSD licenses respectively, and are used by many works, including ours, for research purposes.  We now give details of the datasets we create, building upon CLEVR.
\subsection{\clevrmanip{} Dataset}
This dataset was generated with the help of CLEVR toolkit~\citep{CLEVRDatasetPaper} by using following recipe.
\begin{enumerate}[leftmargin=*]
\item First, we create a source image $I$ and the corresponding scene data by using {\em Blender} \citep{blender} software. 
\item For each source image $I$ created above,  we generate multiple instruction texts $T$'s using its scene data. These are generated using templates, similar to question templates proposed by \cite{CLEVRDatasetPaper}.
\item For each such $(I,T)$ pair, we attach a corresponding symbolic program $P$ (not used by \method{} though) as well as scene data for the corresponding changed image.
\item Finally, for each $(I, T)$ pair, we generate the target gold image $\widetilde{I}^*$ using Blender software and its scene data from the previous step.
\end{enumerate}

Below are some of the important characteristics of the \clevrmanip{} dataset.
\begin{itemize}[leftmargin=*]
\item Each source image $I$ comprises several objects and each object comprises four visual attributes - {\em color, shape, size,} and {\em material}.
\item Each instructions text $T$ comprises one of the following three kinds of manipulation operations - {\em  add, remove,} and {\em change.} 
\item An {\em add} instruction specifies {\em color, shape, size,} and {\em material} of the object that needs to be added. It also specifies a direct (or indirect) relation with one or more existing objects (called reference object(s)). The number of relations that are required to traverse for nailing down the target object is referred to as {\em \# of reasoning hops} and we have allowed instructions with up to {\em $3$-hops reasoning}. We do not generate any $0$-hop instruction for {\em add} due to ambiguity of where to place the object inside the scene. 
\item A {\em change} instruction first specifies zero or more attributes to uniquely identify the object that needs to be changed. It may also specify a direct (or indirect) relation with one or more existing reference objects. Lastly, it specifies the target values of an attribute for the identified object which needs to be changed.
\item A {\em remove} instruction specifies zero or more attributes of the object(s) to be removed. Additionally, it may specify a direct (or indirect) relation with one or more existing reference objects. \end{itemize}

Table \ref{tab:dataset_clevr_manip} captures the fine grained statistics about the \clevrmanip{} dataset. Specifically, it further splits each of the {\em train, validation,} and {\em test} set across the instruction types - {\em add, remove,} and {\em change}. 
    \begin{table*}[ht]
    \centering
    \setlength\tabcolsep{8pt} 
    \renewcommand{\arraystretch}{1.3}
    \setlength\extrarowheight{2pt}
    {
    \begin{tabular}{llrrrrrrrrr}
    \toprule
            \multirow{2}{*}{Operation} & \multirow{2}{*}{Split} & \multirow{2}{*}{\# $(I, T, \widetilde{I}^*)$} && \multicolumn{3}{c}{ \# reasoning hops} && \multicolumn{3}{c}{ \# objects} \\
            \cline{5-7}\cline{9-11}
            &&&& min & mean & max && min & mean & max\\
            \midrule
                    & train & 17827 && 1 & 2.00 & 3 && 3 & 5.51 & 8  \\
            Add     & valid & 4459  && 1 & 2.00 & 3 && 3 & 5.50 & 8  \\
                    & test  & 4464  && 1 & 2.00 & 3 && 3 & 5.45 & 8  \\
    
            \midrule
                   & train & 15999 && 0 & 1.50 & 3 && 3 & 5.50 & 8  \\
            Remove & valid & 5000  && 0 & 1.50 & 3 && 3 & 5.50 & 8  \\
                   & test  & 5000  && 0 & 1.50 & 3 && 3 & 5.48 & 8  \\
    
            \midrule
                   & train & 19990 && 0 & 1.50 & 3 && 3 & 5.45 & 8  \\
            Change & valid & 4996  && 0 & 1.50 & 3 && 3 & 5.56 & 8  \\
                   & test  & 4998  && 0 & 1.50 & 3 && 3 & 5.52 & 8  \\
            
            \bottomrule
            
        \end{tabular}
        }
        \caption{\label{tab:dataset_clevr_manip} Statistics of \clevrmanip{} dataset introduced in this paper.}
    \end{table*}
\subsection{\clevrmaniplarge{} Dataset}

We created another dataset called \clevrmaniplarge{} to test the generalization ability of \method{} on images containing more number of objects than training images. \clevrmaniplarge{} tests the {\em zero-shot transfer} ability of both \method{} and baselines on scenes containing more objects. 

Each image in \clevrmaniplarge{} dataset comprises of $10-13$ objects as opposed to $3-8$ objects in \clevrmanip{} dataset which was used to train \method{}.
The \clevrmaniplarge{} dataset consists of $1K$ unique input images. We have created $3$ instructions for each image resulting in a total of $3K$ instructions. The number of {\em add} instructions is significantly less since there is very little free space available in the scene to add new objects. To create scenes with $12$ and $13$ objects, we made all objects as {\em small size} and the minimum distance between objects was reduced so that all objects could fit in the scene. Table \ref{tab:dataset_clevr_manip_large} captures the statistics about this dataset.
    \begin{table*}[ht]
    \centering
    \setlength\tabcolsep{8pt} 
    \renewcommand{\arraystretch}{1.3}
    \setlength\extrarowheight{2pt}
    {
    \begin{tabular}{lrrccccccc}
    \toprule
            \multirow{2}{*}{Operation} & \multirow{2}{*}{\# $(I, T,  \widetilde{I}^{*})$} && \multicolumn{3}{c}{ \# reasoning hops} && \multicolumn{3}{c}{ \# objects} \\
            \cline{4-6}\cline{8-10}
            &&& min & mean & max && min & mean & max\\
            \midrule
            Add     & 393  && 1 & 2.0 & 3 && 10 & 11.53 & 13  \\
            Remove & 524  && 0 & 1.50 & 3 && 10 & 11.48 & 13  \\
            Change & 2083  && 0 & 1.51 & 3 && 10 & 11.50 & 13   \\
            \bottomrule
        \end{tabular}
        }
        \caption{\label{tab:dataset_clevr_manip_large} Statistics of \clevrmaniplarge{} dataset. }
    \end{table*}
\subsection{Multi-hop Instructions}
In what follows, we have given examples of the instructions that require multi-hop reasoning to nail down the location/object to be manipulated in the image.
\begin{itemize}[leftmargin=*]
\item {\em Remove the tiny green rubber ball.}  {(0-hop)}
\item {\em There is a block right of the tiny green rubber ball, remove it.}  {(1-hop)}
\item {\em Remove the shiny cube left of the block in front of the gray thing.} {(2-hop)}
\item {\em Remove the small thing that is left of the brown matte object behind the tiny cylinder that is behind the big yellow metal block.} {(3-hop)}
\end{itemize}
\section{Model Details}
\label{app:model_details}
\subsection{Semantic Parser}
\subsubsection{Details on Parsing}
We begin by extending the type system of \cite{NSCL} and add \texttt{ConceptSet} because our {\em add} operation takes as input a set of concepts depicting attribute values of the new object being added (refer Table \ref{table_type_system} for the details). Next, in a manner similar to \cite{NSCL}, we use a rule based system for extracting concept words from the input text. We, however, add an extra rule for extracting \texttt{ConceptSet} from the input sentence. Rest of the semantic parsing methodology remains the same as given in  \cite{NSCL}, with the difference being that our training is weakly supervised (refer Section \ref{subsec:neurosim_models} of the main paper).
\subsubsection{Training}
As explained in Section \ref{subsec:neurosim_models} of the main paper, for training with weaker form of supervision, we use an off-policy program search based {\texttt{REINFORCE}} \citep{williams1992simple} algorithm for calculating the exact gradient. For this, we define a set of all possible program templates $\mathbb{P}_t$. For a given input instruction text $T$, we create a set of all possible programs $\{{P}_{T}\}$ from $\mathbb{P}_t$. For e.g. given a template $\{ remove(relate(\cdot, filter(\cdot, scene())))\}$, this is filled in all possible ways, with concepts, conceptSet, attributes and relational concepts extracted from the input sentence to get programs for this particular template. All such programs created using all templates form the set ${P}_{T}$. All ${P}_{T}$ are executed over the scene graph of the input image. A typical program structure in our work is of the form \textit{manip\_op(reasoning())}, where \textit{manip\_op} represents the manipulation operator, for example {\em change, add,} or {\em remove}; and \textit{reasoning()} either selects objects for {\em change} or {\em remove}, or it selects a reference object for adding another object in relation to it. After a hyperparameter search for the reward (refer Section \ref{hyperparams} of the appendix), we assign a reward of +8 if the \textit{reasoning()} part of the program leads to an object being selected for {\em change/remove} instruction or a related object being selected for {\em add} instruction. If no such object is selected, we give a reward of +2. Reward values were decided on the basis of validation set accuracy. We find that with this training strategy, we achieve the validation set accuracy of $95.64 \%$, where this accuracy is calculated based on whether a program lead to an object being selected or not. Note, this is a proxy to the actual accuracy. For finding the actual accuracy, we would need a validation set of (instruction, ground truth output program) pairs, but we do not use this supervised data for training or validation.
\subsection{Manipulation Network}
In what follows, we provide finer details of manipulation network components. 
\paragraph{Change Network:}
As described in Section \ref{subsec:neurosim_models} of the main paper, we have a {\em change neural network} for each attribute. For changing the current attribute value of a given object $o$, we use the following neural network: $\widetilde{o} = g_{a}(o; c_{s_a^*})$, where $s_a^*$ is the desired changed value for the attribute $a$. $\widetilde{o}$ is the new representation of the object. We model $g_{a}(\cdot)$ by a single layer neural network without having any non-linearity. The input dimension of this neural network is $(256+64)$ because we concatenate the object representation $o \in \mathbb{R}^{256}$ with the desired concept representation $d \in \mathbb{R}^{64}$. We pass this concatenated vector through $g_{a}(\cdot)$ to get the revised representation of the object: $\widetilde{o} \in \mathbb{R}^{256}$.

The loss used to train the weights of the change network is a weighted sum of losses \eqref{eq:Lc} to \eqref{eq:Lobjgan} given in the main paper. This leads to the overall loss function given below.
\begin{dmath*}
L_{\text{overall\_change}} = \lambda_1 \; \ell_a + \lambda_2 \; \ell_{\overline{a}} + \lambda_3 \; \ell_{\text{cycle}} + \lambda_4 \; \ell_{\text{consistency}} + \lambda_5 \; \ell_{\text{objGAN}}
\end{dmath*}
where, 
$\ell_{\text{objGAN}}$ above is the modified GAN loss~\citep{goodfellow2014generative}. 
Here $\lambda_1 = 1$, $\lambda_2 = 1/((\text{num\_attrs}-1)*(\text{num\_concepts}))$, $\lambda_3 = \lambda_4 = {10}^{3}$, and $\lambda_5 = 1/(\text{num\_objects})$. Here, $\text{(num\_objects)}$ is the number of objects in input image, $\text{(num\_attrs)}$ is the total number of attributes for each object, and $\text{(num\_concepts)}$ are the total number of concepts in the NSCL~\citep{NSCL} framework.

\begin{figure*}[ht!]
	\centering
	\subfloat[Change operator overview.\label{fig:change_op_fig}]{%
		\includegraphics[width=\textwidth]{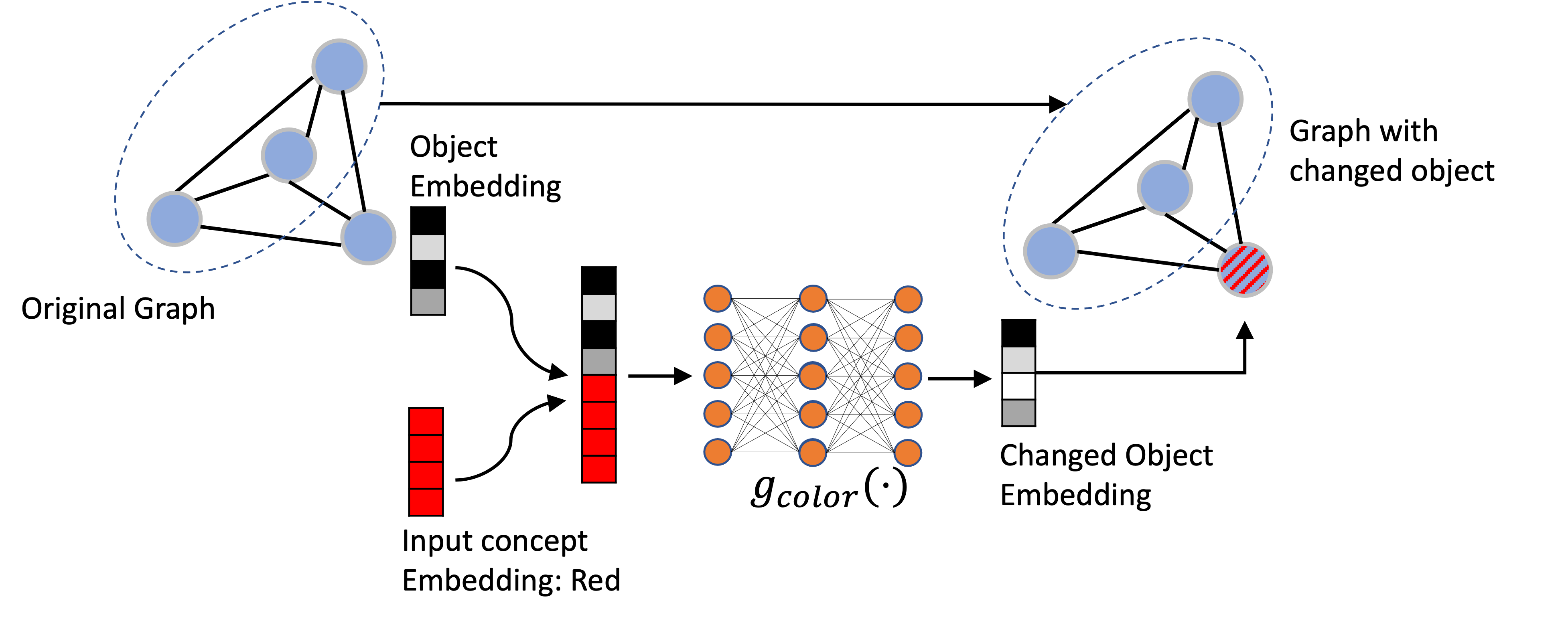}%
	}\par
	\centering
	\subfloat[Add operator overview.\label{fig:add_op_fig}]{%
		\includegraphics[width=\textwidth]{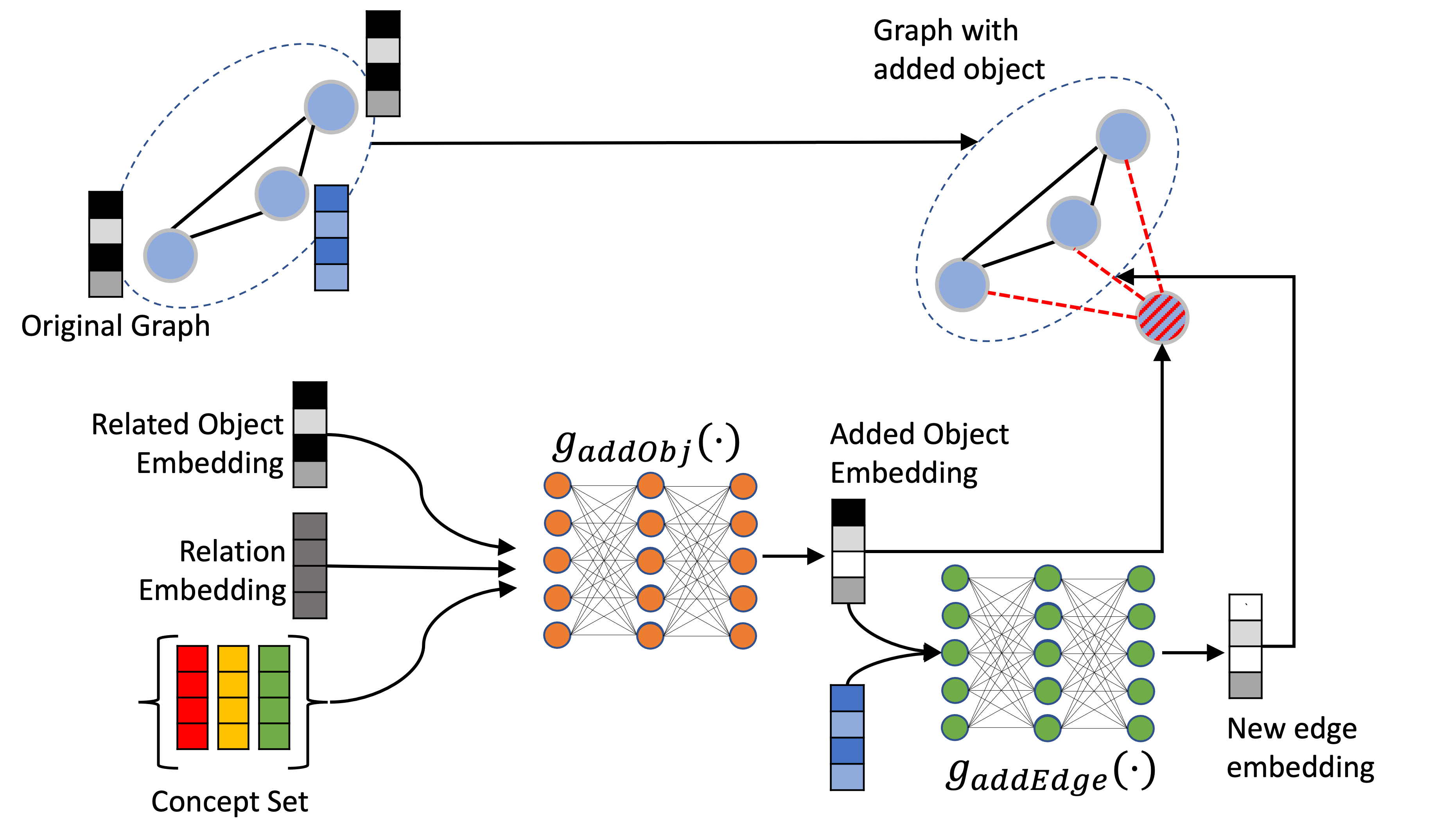}%
	}\par
	\centering
	\subfloat[Remove operator overview.\label{fig:remove_op_fig}]{%
		\includegraphics[width=\textwidth]{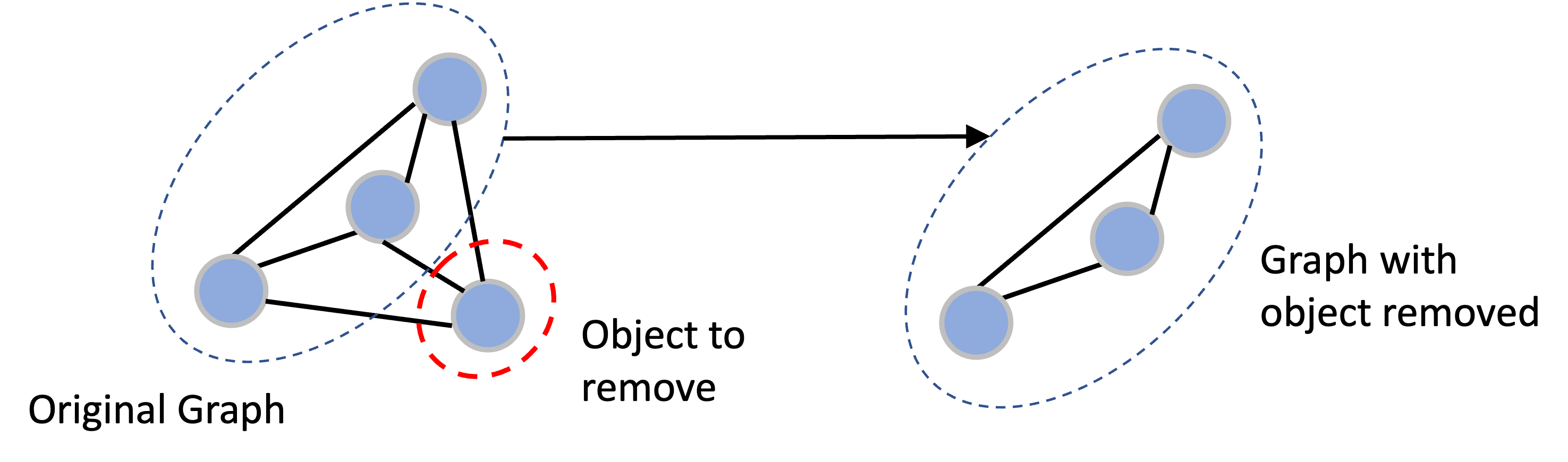}%
	}\par
	\caption{Overview of new operators ({\em change, add} and {\em remove}) added to the DSL.}
	\label{fig:operator_overviews}
	
\end{figure*}

The object discriminator is a neural network with input dimension $256$ and a single $300$ dimensional hidden layer with ReLU activation function. This discriminator is trained using standard GAN objective $\ell_{\text{objGAN}}$.
See Fig \ref{fig:change_op_fig} for an overview of the change operator
\paragraph{Remove Network:} 
The remove network is a symbolic operation as described in Section \ref{subsec:neurosim_models} of the main paper. That is, given an input set of objects, the remove operation deletes the subgraph of the scene graph that contains the nodes corresponding to removed objects and the edges incident on those nodes. See Fig \ref{fig:remove_op_fig} for an overview of the remove operator.
\paragraph{Add Network:} The neural operation in the add operator comprises of predicting the object representation for the newly added object using a function $g_{\text{addObj}}(\cdot)$. This function is modeled as a single layer neural network without any activation. The input to this network is a concatenated vector $[[c_{s_{a_1}}, c_{s_{a_2}}, \cdots, c_{s_{a_k}}], o_{rel}, c_r]$, where $[c_{s_{a_1}}, c_{s_{a_2}}, \cdots, c_{s_{a_k}}]$ represents the concatenation of all the concept vectors of the desired new objects. The vector $o_{rel}$ is the representation of the object with whom the relation (i.e. position) of the new object has been specified and $c_r$ is the concept vector for that relationship. The input dimension of $g_{\text{addObj}}(\cdot)$ is $(k*64 + 256 + 64)$ and the output dimension is $256$.
For predicting representation of newly added edges in the scene graph, we use edge predictor $g_{\text{addEdge}}(\cdot)$. The input to this edge predictor function is the concatenated representation of the objects which are linked by the edge. The input dimension of $g_{\text{addEdge}}(\cdot)$ is $(256+256)$ and the output dimension is $256$.

The loss used to train the {\em add network} weights 
is a weighted sum of losses \eqref{eq:Laddobjconcepts} to \eqref{eq:Ledgegan} along with an object discriminator loss. The overall loss is given by the following expression.
\begin{dmath*}
    L_{\text{overall\_add}} = \lambda_1 \ell_{\text{concepts}} + \lambda_2 \ell_{\text{relation}} + \lambda_3 \ell_{\text{objSup}} + \lambda_4 \ell_{\text{edgeSup}} + \lambda_5
    \ell_{\text{edgeGAN}} + \lambda_6
    \ell_{\text{objGAN}}
\end{dmath*}
where, 
$\ell_{\text{objGAN}}$ and $\ell_{\text{edgeGAN}}$ above denotes the modified GAN loss~\citep{goodfellow2014generative}. 
Here $\lambda_1 = \lambda_2 = 1/(\text{num\_attrs})$, $\lambda_3 = \lambda_4 = {10}^{3}$,  $\lambda_6 = 1/(\text{num\_objects})$.

The object discriminator is a neural network with input dimension as $256$ and a single $300$ dimensional hidden layer with ReLU activation function. This discriminator is trained using the standard GAN objective $\ell_{\text{objGAN}}$. Note, $\ell_{\text{objGAN}}$ has 2 parts -- i) the loss for the generated (fake) object embedding using the \textit{add network}, and ii) the loss for the real objects (all the unchanged object embeddings of the image). The former is unscaled but the latter one is scaled by a factor of $1/(\text{num\_objects})$. 

The edge discriminator is a neural network with input dimension as $(256*3)$ and a single $300$ dimensional hidden layer with ReLU activation function. As input to this discriminator network, we pass the concatenation of the two objects and the edge connecting them. This discriminator is trained using the standard GAN objective $\ell_{\text{edgeGAN}}$. See Fig \ref{fig:add_op_fig} for an overview of the add operator
\section{Additional Results}
\label{supp:experiments}
\subsection{Detailed Performance for Zero-Shot Generalization on Larger Scenes}
Table \ref{tab:combinatorial_add_rem_change} below is a detailed version of the Table \ref{perf_with_hops_and_larger_scenes} in the main paper. This table compares the performance of \method{} with baseline methods TIM-GAN, GeNeVA and IP2P for the zero-shot generalization to larger scenes (with $\ge10$ objects), while the models were trained on images with $3 - 8$ objects. Relative to the main paper's table \ref{perf_with_hops_and_larger_scenes}, this table offers separate performance numbers for each of the {\em add, remove} and {\em change} instructions.

\begin{table*}[ht]
    \setlength\tabcolsep{5pt}
    \renewcommand{\arraystretch}{1.3}
    \setlength\extrarowheight{2pt}
    \centering
    \begin{tabular}{llrrrrrrrrr}
        \toprule[0.7pt]
        \multirow{2}{*}{{Method}}& \multirow{2}{*}{\parbox{0.5cm}{Train Data Size}} && \multicolumn{2}{c}{Add} && \multicolumn{2}{c}{Change} && \multicolumn{2}{c}{Remove}\\
        \cline{4-5} \cline{7-8} \cline{10-11}
        &&& $R1$    & $R3$  && $R1$  & $R3$  && $R1$    & $R3$  \\
        \toprule
        GeNeVA & $54K$        && $0.5$ & $64.6$ && $4.9$ & $69.9$ && $9.0$ & $50.0$ \\
        GeNeVA & $5.4K$       && $0.0$ & $60.1$ && $8.2$ & $69.2$ && $14.3$ & $49.6$ \\
        TIMGAN & $54K$        && $12.5$ & $77.4$ && $73.4$ & $95.2$ && $78.2$ & $92.2$ \\
        TIMGAN & $5.4K$       && $1.0$ & $70.0$ && $32.1$ & $84.4$ && $44.7$ & $74.0$ \\
        IP2P & $54K$        && $38.2$ & $100.0$ && $72.7$ & $100.0$ && $78.6$ & $98.9$ \\
        IP2P & $5.4K$       && $34.1$ & $100.0$ && $68.8$ & $100.0$ && $72.5$ & $96.4$ \\
        \method{} & $5.4K$         && $3.8$ & $46.6$ && $68.2$ & $95.8$ && $90.7$ & $94.3$ \\

        \bottomrule[1.0pt] 
    \end{tabular}
\caption{Detailed performance scores for \method{}, TIM-GAN, GeNeVA and IP2P for zero-shot generalization to larger scenes (with $\ge 10$ objects) from \clevrmaniplarge{} dataset, while models are trained on images with $3 - 8$ objects. Table has separate performance numbers for {\em add, remove,} and {\em change} instructions. Along with each method, we have also written the number of data points from \clevrmanip{} dataset that were used for training. $R1$ and $R3$ correspond to Recall$@1$ and Recall$@3$, respectively.}
\label{tab:combinatorial_add_rem_change}
\end{table*}
\subsection{Image Retrieval Task}
A task that is closely related to the image manipulation task is the task of {\em Text Guided Image Retrieval}, proposed by \cite{Vo_2019_CVPR}. Through this experiment, our is to demonstrate that \method{} is highly effective in solving this task as well. In what follows, we provide details about this task, baselines, evaluation metric, how we adapted \method{} for this task, and finally performance results in Table \ref{tab:image_retrieval_multihop_add_rem_change}. This table is a detailed version of the Table \ref{quantitative_comparisons} in the main paper.
\paragraph{Task Definition:} Given an Image $I$, a text instruction $T$, and a database of images $D$, the task is to retrieve an image from the database that is semantically as close to the ground truth manipulated image as possible. 

Note, for each such $(I, T)$ pair,  some image from the database, say $\widetilde{I} \in D$, is assumed to be the ideal image that should ideally be retrieved at rank-1. This, so called desired gold retrieval image might even be an image which is the ideal manipulated version of the original images $I$ in terms of satisfying the instruction $T$ perfectly. Or, image $\widetilde{I}$ may not be such an ideal manipulated image but it still may be the image in whole corpus $D$ that comes closest to the ideal manipulated image.  

In practice, while measuring the performance of any such system for this task, the gold manipulated image for $(I, T)$ pair is typically inserted into the database $D$ and such an image then serves as the desired gold retrieval image $\widetilde{I}$.
\paragraph{Baselines:} Our baselines includes popular supervised learning systems designed for this task. The first baseline is TIRG proposed by \citet{Vo_2019_CVPR} where they combine image and text to get a joint embedding and train their model in a \textit{supervised} manner using embedding of the desired retrieved image as supervision. For completeness, we also include comparison with other baselines -- {\em Concat, Image-Only}, and {\em Text-Only} -- that were introduced by \citet{Vo_2019_CVPR}.

A recent model proposed by \citet{GEDReward} uses symbolic scene graphs (instead of embeddings) to retrieve images from the database. Motivated by this, we also retrieve images via the scene graph that is generated by the manipulation module of \method{}. However, unlike \citet{GEDReward}, the nodes and edges in our scene graph have associated vectors and make a novel use of them while retrieving.  We do not compare our performance with \cite{GEDReward} since its code is unavailable and we haven't been able to reproduce their numbers on datasets used in their paper. Moreover, \cite{GEDReward} uses full supervision of the desired output image (which is converted to a symbolic scene graph), while we do not. 
\paragraph{Evaluation Metric:} We use Recall$@k$ (and report results for $k=1,3$) for evaluating the performance of text guided image retrieval algorithms which is  standard in the literature.
\paragraph{Retrieval using Scene Graphs:} We use the scene graph generated by \method{} as the latent representation to retrieve images from the database. We introduce a novel yet simple method to retrieve images using scene graph representation. For converting an image into the scene graph, we use the visual representation network of \method{}. Given the scene graph $G$ for the input image $I$ and the manipulation instruction text $T$, \method{} converts the scene graph into the changed scene graph $G_{\widetilde I}$, as described in Section \ref{app:model_details} in Appendix. Now, we use this graph $G_{\widetilde I}$ as a query to retrieve images from the database $D$. For retrieval, we use the novel graph edit distance (GED) between $G_{\widetilde I}$ and the scene graph representation of the database images. The scene graph for each database image is also obtained using the visual representation network of \method{}. The graph edit distance is given below.
{
\begin{equation*}  
    \resizebox{0.45\textwidth}{!}{$GED(G_{\widetilde{I}},G_{D}) = 
         \begin{cases}
           \infty &\quad |{N}_{\widetilde{I}}|  \neq |{N}_{\widetilde{D}}| \\
           \min_{\pi \in \Pi}{\sum_{\forall i \in \{1, \cdots, |{N}_{\widetilde{I}}|\}} c(n_i,y_i)} &\quad\text{otherwise.}
         \end{cases}$
    }
\end{equation*}
}
where, $G_{\widetilde{I}}=(N_{\widetilde{I}}, V_{\widetilde{I}})$ and  $G_{D}=(N_{D}, V_{D})$.  $n_i$ and $y_i$ are the node embeddings of the query graph $G_{\widetilde{I}}$ and scene graph  $G_{D}$ of an image from the database. $c(a,b)$ is the cosine similarities between embeddings $a$ and $b$.
This GED is much simpler than that defined in \cite{GEDReward}, since it does not need any hand designed cost for {\em change, removal,} or {\em addition} of nodes, or different attribute values. It can simply rely on the cosine similarities between node embeddings. We use the \textit{Hungarian algorithm}~\citep{Hungarian} for calculating the optimal matching $\pi$ of the nodes, among all possible matching $\Pi$. We use the negative of the cosine similarity scores between nodes to create the cost matrix for the Hungarian algorithm to process. This simple yet highly effective approach (See Table \ref{quantitative_comparisons} in the main paper and Table \ref{tab:image_retrieval_multihop_add_rem_change} in the appendix), can be improved by more sophisticated techniques that include distance between edge embeddings and including notion of subgraphs in the GED. We leave this as future work.
This result shows that our manipulation network edits the scene graph in a desirable manner, as per the input instruction.

\begin{table*}[ht]
    \setlength\tabcolsep{5pt}
    \renewcommand{\arraystretch}{1.3}
    \setlength\extrarowheight{2pt}
    \centering
    \begin{tabular}{llrrrrrrrrrrrrrr}
        \toprule[0.7pt]
        \multirow{2}{*}{{Method}}& \multirow{2}{*}{\parbox{0.5cm}{Train Data Size}}&& \multicolumn{3}{c}{Add} && \multicolumn{4}{c}{Change} && \multicolumn{4}{c}{Remove}\\
        \cline{4-6} \cline{8-11} \cline{13-16}
        &&& $1$    & $2$    & $3$ && $0$    & $1$    & $2$    & $3$ && $0$    & $1$    & $2$    & $3$\\
        \toprule
        Text-Only & $54K$         && $0.4$ & $0.3$ & $0.3$ && $0.1$ & $0.0$ & $0.1$ & $0.1$ && $0.1$ & $0.3$ & $0.3$ & $0.0$ \\
        Image-Only & $54K$        && $35.3$ & $33.4$ & $32.3$ && $20.1$ & $23.2$ & $16.9$ & $19.8$ && $46.3$ & $41.3$ & $53.1$ & $57.8$ \\
        Concat & $54K$            && $36.3$ & $33.3$ & $31.8$ && $37.3$ & $40.4$ & $34.2$ & $37.9$ && $41.8$ & $41.0$ & $50.0$ & $55.0$ \\
        TIRG & $54K$              && $35.6$ & $31.8$ & $33.5$ && $22.0$ & $25.1$ & $18.8$ & $22.0$ && $46.6$ & $42.7$ & $52.5$ & $56.1$ \\
        \method{} & $5.4K$        && $96.2$ & $95.3$ & $95.3$ && $83.3$ & $82.9$ & $81.3$ & $78.7$ && $79.6$ & $77.4$ & $86.4$ & $82.2$ \\
        \bottomrule[1.0pt] 
    \end{tabular}
\caption{Performance scores (Recall@1) on the Image Retrieval task, comparing \method{} with TIM-GAN and GeNeVA with increase in reasoning hops, for {\em add, remove,} and {\em change} instructions. Along with each method, number of data points from \clevrmanip{} used for training are written.}
\label{tab:image_retrieval_multihop_add_rem_change}
\end{table*}
\begin{table*}[ht]
    \setlength\tabcolsep{4.3pt}
    \renewcommand{\arraystretch}{1.3}
    \setlength\extrarowheight{2pt}
    \centering
    \begin{tabular}{lllrrrrrrrrrrrrrrr}
    \toprule[1.0pt]
    \multirow{3}{*}{{Method}} 
    && \multirow{3}{*}{{Instruction}} 
    && \multicolumn{2}{c}{$\beta =0.054$} 
    && \multicolumn{2}{c}{$\beta =0.07$}
    && \multicolumn{2}{c}{$\beta =0.1$} 
    && \multicolumn{2}{c}{$\beta =0.2$}  
    && \multicolumn{2}{c}{$\beta =0.54$} \\

    \cline{5-6} \cline{8-9} \cline{11-12} \cline{14-15} \cline{17-18} &&
    && {$R1$} & {$R3$} && {$R1$} & {$R3$} && {$R1$} & {$R3$} && {$R1$} & {$R3$} && {$R1$} & {$R3$} \\ \midrule
    \multirow{3}{*}{\small GeNeVA}
    && add      && $0.0$ & $57.3$ && --  & --  && --  & --  && --  & --  && $0.7$ & $63.6$ \\ 
    && change   && $5.9$ & $36.3$ && --  & --  && --  & --  && --  & --  && $4.1$ & $39.4$ \\
    && remove   && $13.2$ & $82.3$ && --  & --  && --  & --  && --  & --  && $8.7$ & $89.3$ \\
    \midrule
    \multirow{3}{*}{\small TIM-GAN}
    && add      && $1.9$ & $70.7$ && $4.9$ & $74.0$ && $8.6$ & $76.7$ && $10.3$ & $77.1$ && $13.1$ & $78.6$\\
    && change   && $41.0$ & $72.1$ && $42.9$ & $73.5$ && $49.8$ & $77.3$ && $62.5$ & $84.2$ && $78.3$ & $92.3$\\
    && remove   && $49.6$ & $79.5$ && $47.0$ & $91.9$ && $53.9$ & $93.1$ && $65.3$ & $96.8$ && $78.0$ & $98.5$\\
    \midrule
    \multirow{3}{*}{\small IP2P}
    && add      && $0.7$ & $70.5$ && --  & --  && --  & --  && --  & --  && $5.4$ & $78.6$ \\ 
    && change   && $54.9$ & $72.8$ && --  & --  && --  & --  && --  & --  && $61.5$ & $78.9$ \\
    && remove   && $62.0$ & $87.2$ && --  & --  && --  & --  && --  & --  && $76.0$ & $96.3$ \\
    \midrule
    \multirow{3}{*}{\small \method{}}
    && add      && $4.9$ & $30.9$ && $6.4$ & $34.8$ && $ 5.7$ & $34.7$ && $5.9$ & $38.9$ && $5.6$ & $35.0$  \\
    && change   && $57.2$ & $79.4$ && $57.3$ & $79.3$ && $57.2$ & $79.3$ && $57.2$ & $79.4$ && $57.1$ & $79.3$  \\
    && remove   && $69.6$ & $82.5$ && $69.5$ & $82.5$ && $69.5$ & $82.6$ && $69.5$ & $82.5$ && $69.6$ & $82.5$  \\
    \bottomrule[1.0pt]
\end{tabular}
\caption{Detailed performance comparison of \method{} with TIM-GAN~\citep{TIMGAN}, GeNeVA~\citep{GeNeVa} and IP2P~\citep{brooks2022instructpix2pix} with varying $\beta$ levels, split across add, remove and change instructions. The '-' entries for GeNeVA and IP2P were not computed due to excessive training  time (inference time as well in case of IP2P); Geneva's performance is abysmal even when using full data. TIM-GAN does the best among baselines in terms of its recall score at $\beta=0.54$. We always use $100K$ VQA examples (5K Images, 20 questions per image) for our weakly supervised training. $R1$ and $R3$ correspond to Recall@1 and 3, respectively. For {Recall}, higher score is better.}
\label{quantitative_comparisons_combined_supp}
\end{table*}
\subsection{Detailed Multi-hop Reasoning Performance}
\begin{table*}[ht]
    \setlength\tabcolsep{4pt}
    \renewcommand{\arraystretch}{1.3}
    \setlength\extrarowheight{2pt}
    \centering
    \begin{tabular}{llrrrrrrrrrrrrrr}
        \toprule[0.7pt]
        \multirow{2}{*}{{Method}} &\multirow{2}{*}{\parbox{0.8cm}{Train Data Size}}&& \multicolumn{3}{c}{Add} && \multicolumn{4}{c}{Change} && \multicolumn{4}{c}{Remove}\\
        \cline{4-6} \cline{8-11} \cline{13-16}
        &&& $1$    & $2$    & $3$ && $0$    & $1$    & $2$    & $3$ && $0$    & $1$    & $2$    & $3$\\
        \toprule
        GeNeVA    & $54K$  && $1.1$ & $0.5$ & $0.5$ && $3.6$ & $4.2$ & $4.7$ & $3.9$ && $9.0$ & $8.0$ & $8.3$ & $9.4$ \\
        GeNeVA & $5.4K$       && $0.0$ & $0.0$ & $0.0$ && $4.7$ & $7.1$ & $5.6$ & $6.1$ && $12.3$ & $11.3$ & $15.5$ & $13.5$ \\
        TIM-GAN & $54K$       && $7.6$ & $16.1$ & $15.7$ && $85.8$ & $74.1$ & $78.0$ & $75.4$ && $82.2$ & $68.3$ & $81.9$ & $79.7$ \\ 
        TIM-GAN & $5.4K$      && $1.4$ & $2.3$ & $1.9$ && $54.5$ & $36.4$ & $38.7$ & $34.5$ && $58.3$ & $40.9$ & $50.9$ & $48.2$ \\
        IP2P & $54K$       && $6.6$ & $5.5$ & $4.2$ && $67.9$ & $60.6$ & $59.1$ & $58.5$ && $77.1$ & $71.9$ & $78.1$ & $76.8$ \\ 
        IP2P & $5.4K$      && $0.5$ & $0.7$ & $0.9$ && $64.0$ & $54.8$ & $50.6$ & $50.3$ && $74.5$ & $55.3$ & $60.3$ & $58.0$ \\
        \method{} & $5.4K$    && $4.6$ & $5.0$ & $5.1$ && $59.5$ & $57.9$ & $55.8$ & $55.7$ && $69.6$ & $66.6$ & $71.8$ & $70.4$ \\
        \bottomrule[1.0pt] 
    \end{tabular}
\caption{Performance scores (Recall@1) for \method{} with TIM-GAN, GeNeVA and IP2P with increase in reasoning hops, for {\em add, remove,} and {\em change} instructions. Along with each method, number of data points from \clevrmanip{} used for training are written.}
\label{tab:multihop_add_rem_change}
\end{table*}

Table \ref{tab:multihop_add_rem_change} below provides a detailed split of the performance numbers reported in Table \ref{perf_with_hops_and_larger_scenes} of the main paper across i) number of hops ($0 - 3$ hops) and ii) type of instructions {\em (add/remove/change)}. We observe that for {\em change} and {\em remove} instructions, \method{} improves over TIM-GAN, GeNeVA and IP2P trained on $5.4K$ \clevrmanip{} data points by a significant margin ($\sim 20\%$ on $3$-hop {\em change/remove} instructions). However, \method{} lags behind TIM-GAN when the entire \clevrmanip{} labeled data is used to train TIM-GAN. We also observe that all the models perform poorly on the {\em add} instructions, as compared to {\em change} and {\em remove} instructions.

\subsection{Detailed Performance for Different Cost Ratios $\beta$}
Table \ref{quantitative_comparisons_combined} in Section \ref{sec_experiments} of the main paper showed the performance of \method{} compared with TIM-GAN and GeNeVA for various values of $\beta$, where $\beta$ is the ratio of the number of annotated (with output image supervision) image manipulation examples required by the supervised baselines, to the number of annotated VQA examples required to train \method{}. In Table \ref{quantitative_comparisons_combined_supp}, we show a detailed split of the performance, for the {\em add, change,} and {\em remove} operators, across the same values of $\beta$ as taken before. 

We find that for the {\em change} operator, \method{} performs better than TIM-GAN by a margin of $\sim 8\%$ (considering Recall@1) for $\beta \leq 0.1$. For the {\em remove} operator, \method{} performs better than TIM-GAN by a margin of $\sim 4\%$ (considering Recall@1) for $\beta \leq 0.2$. Overall, \method{} performs similar
to TIM-GAN, for $\beta = 0.2$, for {\em remove} and {\em change} operators. All models perform poorly on the {\em add} operator as compared to the {\em change} and {\em remove} operators. We find that having full output image supervision allows TIM-GAN to reconstruct (copy) the unchanged objects from the input to the output for all the operators. This results in a higher recall in general but its effect is most pronounced in the Recall$@3$. \method{}, on the other hand, suffers from rendering errors which makes the overall recall score (especially Recall$@3$) lower. We believe that improving image rendering quality would significantly improve the performance of \method{} and we leave this as future work. 

\begin{figure*}[h!]
	\centering
        \vspace{-0.3cm}
	\includegraphics[page=1,width=\textwidth]{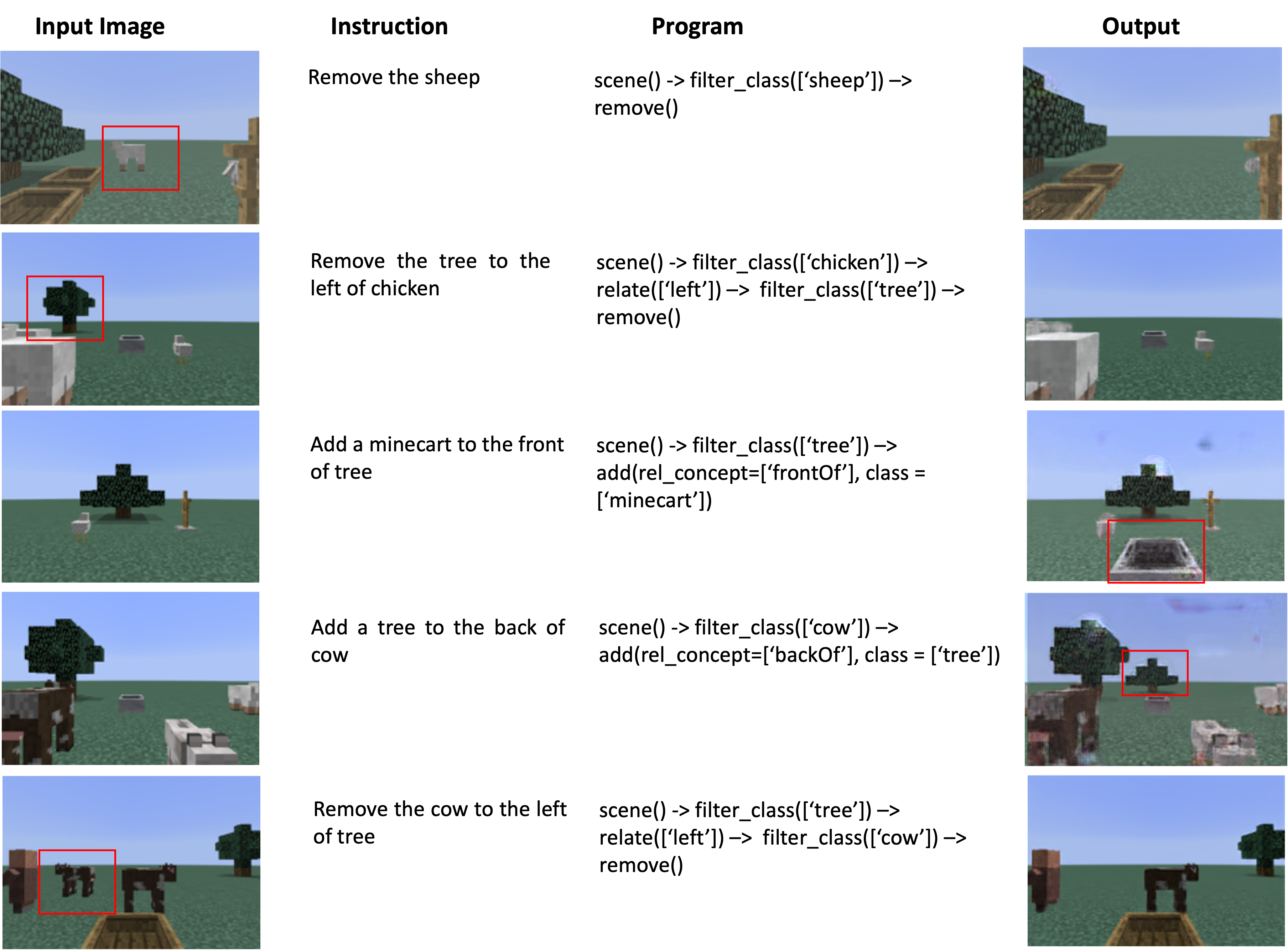}
	\caption{ Results for addition and removal of objects from images of the minecraft dataset}
	\label{fig:minecraft_add_remove}
\end{figure*}

{
\subsection{Results on Datasets from different domains}

\subsubsection{Minecraft Dataset}

\textbf{Dataset Creation:}
We create a new dataset having (Image, instruction) by building over the Minecraft dataset used in \cite{NSVQA}. Specifically, we create zero and one hop remove instructions and one hop add instructions similar to the creation of \clevrmanip{}.
This dataset contains scenes and objects from the Minecraft video game and is used in prior works for testing Neuro-Symbolic VQA systems like NSCL \cite{NSCL} and NS-VQA \cite{NSVQA}. The setting of the Minecraft worlds dataset is significantly different from CLEVR in terms of concepts and attributes of objects and visual appearance.

\textbf{Experiment:}
We use the above dataset for testing the addition and removal of objects using NeuroSIM (See Fig \ref{fig:minecraft_add_remove}). We train NeuroSIM's decoder to generate images from scene graphs of the minecraft dataset. We assume access to a parser that gives us programs for an instruction.
For removal, we use the same remove network as described above, while for addition, we assume access to the features of object to be added, which is added to the scene graph of the image and the decoder decodes the final image. See Figure \ref{fig:minecraft_add_remove} for a set of successful examples on the Minecraft dataset.
We see that using our method, one can add and remove objects from the scene successfully, without using any output image as supervision during training.
Though we have assumed the availability of a parser in the above set-up, training it jointly with other modules should be straightforward, and can be achieved using our general approach described in Section~\ref{section:modeldetails} of the main paper. 
{
\section{End-to-end Training}
\label{supp:end2end}
The main objective of this work is to make use of weakly supervised VQA data for the image manipulation task without using output image supervision. But a natural extension of our work is to use output image supervision as well, to improve the performance of \method{}.
We devised an experiment to compare how much performance boost can be obtained by utilizing ground truth output (manipulated) images as the supervision for different modules of \method{}. This experiment demonstrates the value of end-to-end training for \method{} and how it can exploit the supervised data. We refer to this variant as \method{}(e2e). We begin with a pre-trained \method{} model trained with VQA annotations and then fine-tune it using supervised manipulation data.
The detailed results are given in Table~\ref{tab:NeuroSIM_e2e_performance}. This experiment demonstrates that with a small amount of supervised data, the performance of \method{} can be significantly improved (e.g., more than $9$ points increase for the change instruction with only $5.4K$ supervision examples)

\begin{table*}[!ht]
\setlength\tabcolsep{10pt}
\renewcommand{\arraystretch}{1.3}
\setlength\extrarowheight{2pt}
\centering
\begin{tabular}{llrrrrrrrr}
\toprule[0.7pt]
\multirow{2}{*}{Instruction} &
  \multirow{2}{*}{Model} & \multicolumn{5}{c}{\# of \clevrmanip{} examples used for training}\\
  \cline{3-7}
  && 5.4K & 7K & 10K & 20K & 54K \\
  \toprule

\multirow{4}{*}{Add}    & GeNeVA        & 0.0             & -             & -             & -             & 0.7           \\
                        & TIM-GAN       & 1.9           & 4.9           & 8.6           & 10.3          & 13.1 \\
                        &IP2P        & 0.7           & -             & -             & -             & 5.4\\
                        & \method{}      & 4.9           & 6.4           & 5.7           & 5.9           & 5.6           \\
                        & \method{}(e2e) & 8.8  & 8.9  & 9.2  & 10.5 & 10.6          \\  \midrule
                        
\multirow{4}{*}{Change} & GeNeVA        & 5.9           & -             & -             & -             & 4.1\\
                        & TIM-GAN       & 41.0            & 42.9          & 49.8          & 62.5          & 78.3 \\
                        & IP2P        & 54.9           & -             & -             & -             & 61.5\\
                        & \method{}      & 57.2          & 57.3          & 57.2          & 57.2          & 57.1          \\
                        & \method{}(e2e) & 66.2 & 66.3 & 66.6 & 67.4 & 69.6          \\ 
                         \midrule

\multirow{4}{*}{Remove} & GeNeVA        & 13.2          & -             & -             & -             & 8.7           \\
                        & TIM-GAN       & 49.6          & 47.0            & 53.9          & 65.3          & 78.0   \\
                        &IP2P        & 62.0          & -             & -             & -             & 76.0\\
                        & \method{}      & 69.6 & 69.5 & 69.5 & 69.5 & 69.6          \\
                        & \method{}(e2e) & 69.6 & 69.5 & 69.5 & 69.5 & 69.6          \\   \bottomrule
\end{tabular}
\caption{Performance comparison of \method{}(e2e) with baselines using Recall@1. \method{}(e2e) refers to \method{} trained end-to-end by utilizing ground truth manipulated images as the supervision for \method{} modules. }
\label{tab:NeuroSIM_e2e_performance}
\end{table*}
Given the significant increase in performance of \method{} when using supervised data, we also test it’s generalization capability (Analogous to Section \ref{expt:multi_hop}, \ref{results:combgen}), and quality of scene graph retrieval (Analogous to Section \ref{results:scene_graph_quality}
).

From Table \ref{tab:zero_shot_generalization_extension}, we see that \method{}(e2e) shows improved zero-shot generalization to larger scenes. Even when trained on just 5.4k CIM-NLI data, \method{}(e2e) improves over TIM-GAN-54k by 3.9 R@1 points. A 5.3 point improvement over TIM-GAN is observed when full CIM-NLI data is used.

\begin{table}[!ht]
\centering
\setlength\tabcolsep{5pt}
\renewcommand{\arraystretch}{1.3}
\setlength\extrarowheight{2pt}
\begin{tabular}{llrrrr}
\toprule[0.7pt]
\multirow{2}{*}{Model}  & \multirow{2}{*}{\parbox{0.5cm}{Train Data Size}}              & \multirow{2}{*}{R1}   & \multirow{2}{*}{R3}   \\ \\
\toprule
TIM-GAN  & 5.4K       & 30.2 & 80.7 \\
TIM-GAN   & 54K        & 66.3 & 92.4 \\
\method{}   & 5.4K      & 63.7 & 89.1 \\
\method{}(e2e)   & 5.4K & 70.2 & 92.6 \\
\method{}(e2e)  & 54K  & 71.6 & 91.7 \\         
\bottomrule[1.0pt] 

\end{tabular}
\caption{Zero-shot generalization to larger scenes (Extension of Table \ref{perf_with_hops_and_larger_scenes} of main paper).}
\label{tab:zero_shot_generalization_extension}
\end{table}

Next, we measure drop in performance with increasing reasoning hops. From Table \ref{tab:perf_drop_with_increased_hop}, we see that \method{}(e2e) achieves the lowest drop when compared to TIM-GAN. \method{}(e2e) improves over weakly supervised \method{} baseline by 6.6 R@1 points.
\begin{table*}[ht]
    \setlength\tabcolsep{5pt}
    \renewcommand{\arraystretch}{1.3}
    \setlength\extrarowheight{2pt}
    \centering
    \begin{tabular}{llrrrrr}
        \toprule[0.7pt]
        \multirow{2}{*}{Method}& \multirow{2}{*}{\parbox{0.5cm}{Train Data Size}} && \multicolumn{3}{c}{Hops} \\ 
        \cline{4-6}
        &&& $ZH$    & $MH$ &  $\triangle$\\
        \toprule
        TIM-GAN & 5.4K        && 56.4 & 41.6 & -14.8 \\
        TIM-GAN & 54K       && 84.0 & 76.2  & -7.8\\
        \method{} & 5.4K        && 64.5 & 63.0  & -1.5 \\
        \method{}(e2e) & 5.4K       && 69.4 & 67.3  & -2.1 \\
        \method{}(e2e) & 54K         && 71.1 & 69.6  & -1.5 \\

        \bottomrule[1.0pt] 
    \end{tabular}
\caption{Performance with increasing reasoning hops (Extension of Table \ref{perf_with_hops_and_larger_scenes} of main paper).}
\label{tab:perf_drop_with_increased_hop}
\end{table*}

Finally, we measure the quality of scene graphs via retrieval. From Table \ref{tab:scene_graph_quality_extension}, we see that supervised training significantly improves the scene graph quality, thus improving retrieval performance. Supervised training improves retrieval by 7.3 R@1 points over weakly supervised \method{} baseline.
\begin{table}[ht]
\centering
\setlength\tabcolsep{5pt}
\renewcommand{\arraystretch}{1.3}
\setlength\extrarowheight{2pt}
\begin{tabular}{lrr}
\toprule
\multicolumn{1}{l}{Model} & \multicolumn{1}{r}{R1} & \multicolumn{1}{r}{R3} \\ 
\toprule
Text-Only     & 0.2           &  0.4            \\
Image-Only    &  34.1           &  83.6           \\
Concat        &  39.5           &  86.9           \\
TIRG          &  34.8           &  84.6           \\
\method{}      &  85.8           &  92.9           \\
\method{}(e2e) &  93.1  &  96.7  \\ \hline
\end{tabular}
\caption{Quality of scene graph measured via retrieval (Extension of Table \ref{quantitative_comparisons} of main paper)}
\label{tab:scene_graph_quality_extension}
\end{table}
These findings suggest that \method{}(e2e) significantly outperforms other supervised approaches in almost all settings. One can fine-tune the image decoder and the visual representation network to further enhance the findings, which should greatly enhance the outcomes. 

}
}

\section{LLMs as few-shot parser}
\label{llm_few_shot_parser}
\begin{table}[!ht]
\centering
\setlength\tabcolsep{3pt}
\renewcommand{\arraystretch}{1.3}
\setlength\extrarowheight{2pt}
\begin{tabular}{lrrrrrrr}
\toprule[0.7pt]
\multirow{2}{*}{\parbox{1.5cm}{Instruction type}} & \multicolumn{4}{c}{Hops} & \multirow{2}{*}{Total} \\
\cline{2-5}
& $0$ & $1$ & $2$  & $3$  \\
\toprule
Add & NA & 18.9 & 30.5 & 37.5 & 29.5 \\
Remove & 91.5 & 85.1 & 80.0 & 84.9 & 85.6 \\
Change & 70.7 & 81.7 & 76.2 & 70.2 & 74.6 \\
Total & 81.5 & 55.6 & 55.9 & 58.1 & 60.5 \\
\bottomrule[1.0pt]
\end{tabular}
\captionsetup{justification=raggedright,singlelinecheck=false}
\caption{Few-shot parsing results of GPT-4}
\label{tab:few_shot_gpt4_results}
\end{table}

We also tested the semantic parsing ability of  Large Language Models (LLMs), specifically GPT-4 for our task. The task of semantic parsing is given manipulation instruction text in natural language, generated the symbolic program by parsing the input text. To provide GPT-4 with context, we designed an extensive prompt that begins with our DSL followed by six different in-context examples representing various instruction types for few-shot learning. This prompt is then followed with the instruction text that we want to parse. We tested GPT-4 on a randomly sampled subset of our test dataset. For evaluation, we measured the accuracy of semantic parsing using an exact match between the generated symbolic program and the ground-truth symbolic program.

The detailed results are given in Table~\ref{tab:few_shot_gpt4_results}. Interestingly, we observed that GPT-4 performed poorly on Add instructions, achieving less than $10\%$ of parsing accuracy. To address this, we prompted GPT-4 separately with additional few-shot examples for Add instructions, which led to the results displayed in the table. Even with the additional guidance, Add instructions remained significantly lower in accuracy compared to other instruction types. This analysis demonstrates that our reinforcement learning-based instruction parser outperforms GPT-4, at least on this dataset. It also highlights the need for more careful prompt engineering before LLMs like GPT-4 can be readily applied in our specific setting.

\section{Computational Resources}
\label{computational_resources}
We trained all our models and baselines on $1$ Nvidia Volta V100 GPU with 32GB memory and 512GB system RAM except IP2P which was trained on 8-A100 80 GB GPUs. Our image decoder training takes about $4$ days of training time. Training of the VQA task takes $5 - 7$ days of training time and training the Manipulation networks take $4 - 5$ hours of training time.
\section{Hyperparameters and Validation Accuracies}
\label{hyperparams}
\subsection{Training for VQA Task}
The hyperparameters for the VQA task are kept same as default values coming from the  prior work~\citep{NSCL}. We refer the readers to \cite{NSCL} for more details. We obtained a question answering accuracy of $99.3\%$ after training on the VQA task.
\subsection{Training Semantic Parser}
\label{app:sem_parser_training}
The semantic parser is trained to parse instructions. Learning of this module happens using the \texttt{REINFORCE} algorithm as described in Section \ref{app:model_details} of this appendix. During \texttt{REINFORCE} algorithm, we search for positive rewards from the set $\{7,8,10\}$, and negative rewards from the set $\{0, 2, 3\}$. We finally choose a positive reward of $8$ and negative reward of $2$. For making this decision, we first train the semantic parser for $20$ epochs and then calculate its accuracy by running it on the quantized scenes from the validation set. For a particular output program, we say it is correct if it leads to an object being selected (see Section \ref{app:model_details} of the appendix for more information) and this is how the accuracy of the semantic parser is calculated. This accuracy is a proxy for the real accuracy. An alternative is to use annotated ground truth programs for calculating accuracy and then selecting hyperparameters. However, we do not use ground truth programs. All other hyperparameters are kept the same as used by \cite{NSCL} to train the parser on VQA task. We obtain a validation accuracy of $95.64 \%$ after training the semantic parser for manipulation instructions.


\subsection{Training Manipulation Networks}
The architecture details of the manipulation network are present in Section \ref{app:model_details} of this appendix. We use batch size of $32$, learning rate of ${10}^{-3}$, and optimize using AdamW~\citep{adamw} with weight decay of ${10}^{-4}$.
Rest of the hyperparameters are kept the same as used in \cite{NSCL}. During training, at every $5^{th}$ epochs, we calculate the manipulation accuracy by using the query networks that were trained while training the \method{} on VQA data. This serves as a proxy to the validation accuracy. 
\begin{itemize}[leftmargin=*]
\item For the {\em change} network training, we use the query accuracy of whether the attribute that was supposed to change for a particular object, has changed correctly or not. Also, whether any other attribute has changed or not.
\item For the {\em add} network training, we use the query accuracy of whether the attributes of the added object are correct or not. Also, whether the added object is in a correct relation with reference object or not. 
\end{itemize}
We obtained a validation accuracy (based on querying) of $95.9\%$ for the {\em add} network and an accuracy of $99.1\%$ for the {\em change} network.
\subsection{Image Decoder Training}
The architecture of the image decoder is similar to \cite{Sg2Im} but our input scene graph (having embeddings for nodes and edges) is directly processed by the graph neural network. We use a batch size of $16$, learning rate of ${10}^{-5}$, and optimize using Adam~\citep{adam} optimizer. The rest of the hyperparameters are same as \cite{Sg2Im}. We train the image decoder for a fixed set of $1000K$ iterations.
\section{Qualitative Analysis}
\label{app:qualitative}
\begin{figure*}[h!]
	\centering
	\includegraphics[page=2,width=\textwidth]{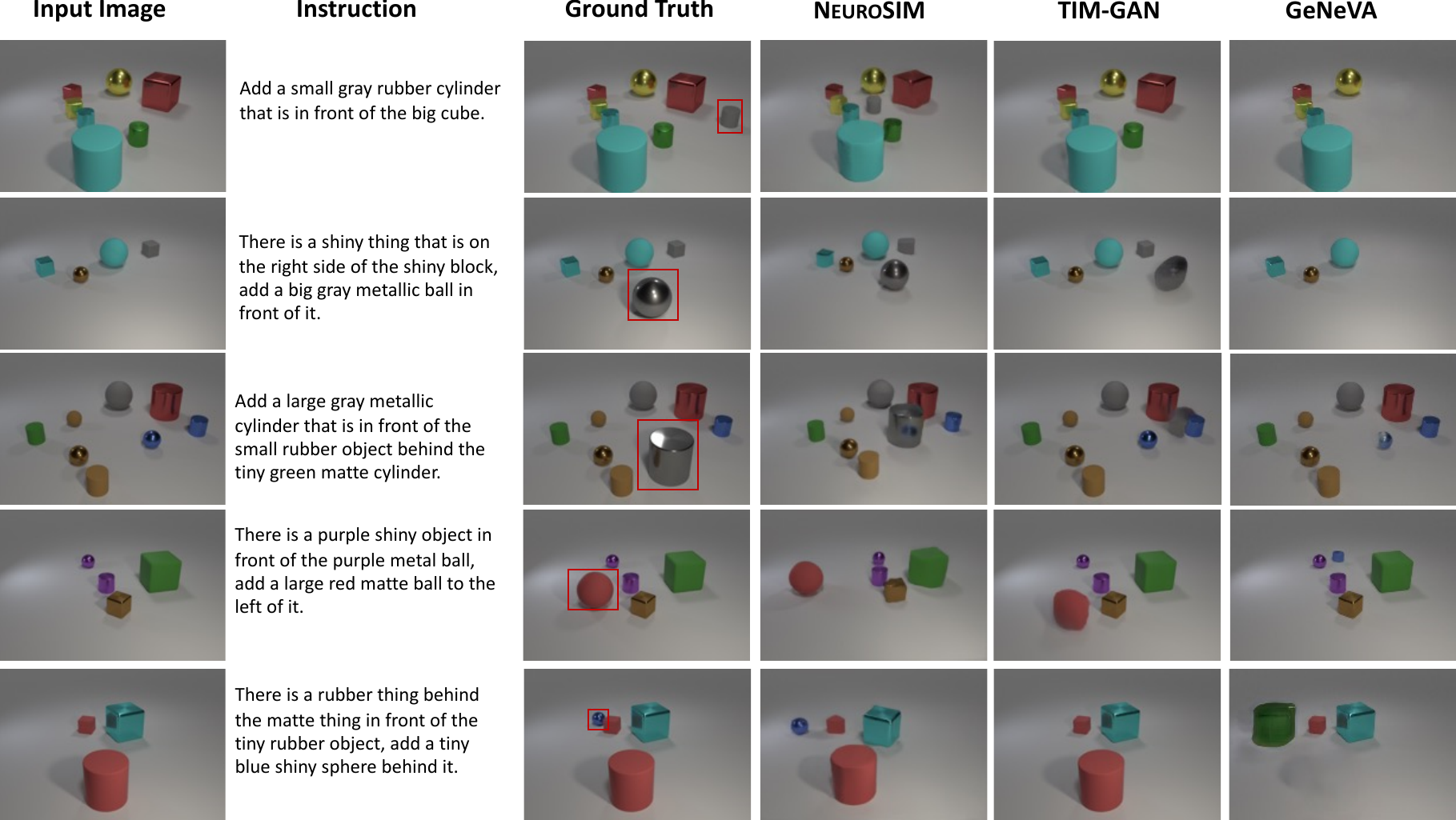}
	\caption{Visual comparison of \method{} with TIM-GAN and GeNeVA for the \textit{add} operator. The red bounding boxes in the ground truth output image indicate the objects required to add to the input image.}
	\label{fig:app_visual_comparisons_add}
\end{figure*}
\begin{figure*}[h!]
	\centering
	\includegraphics[page=1,width=\textwidth]{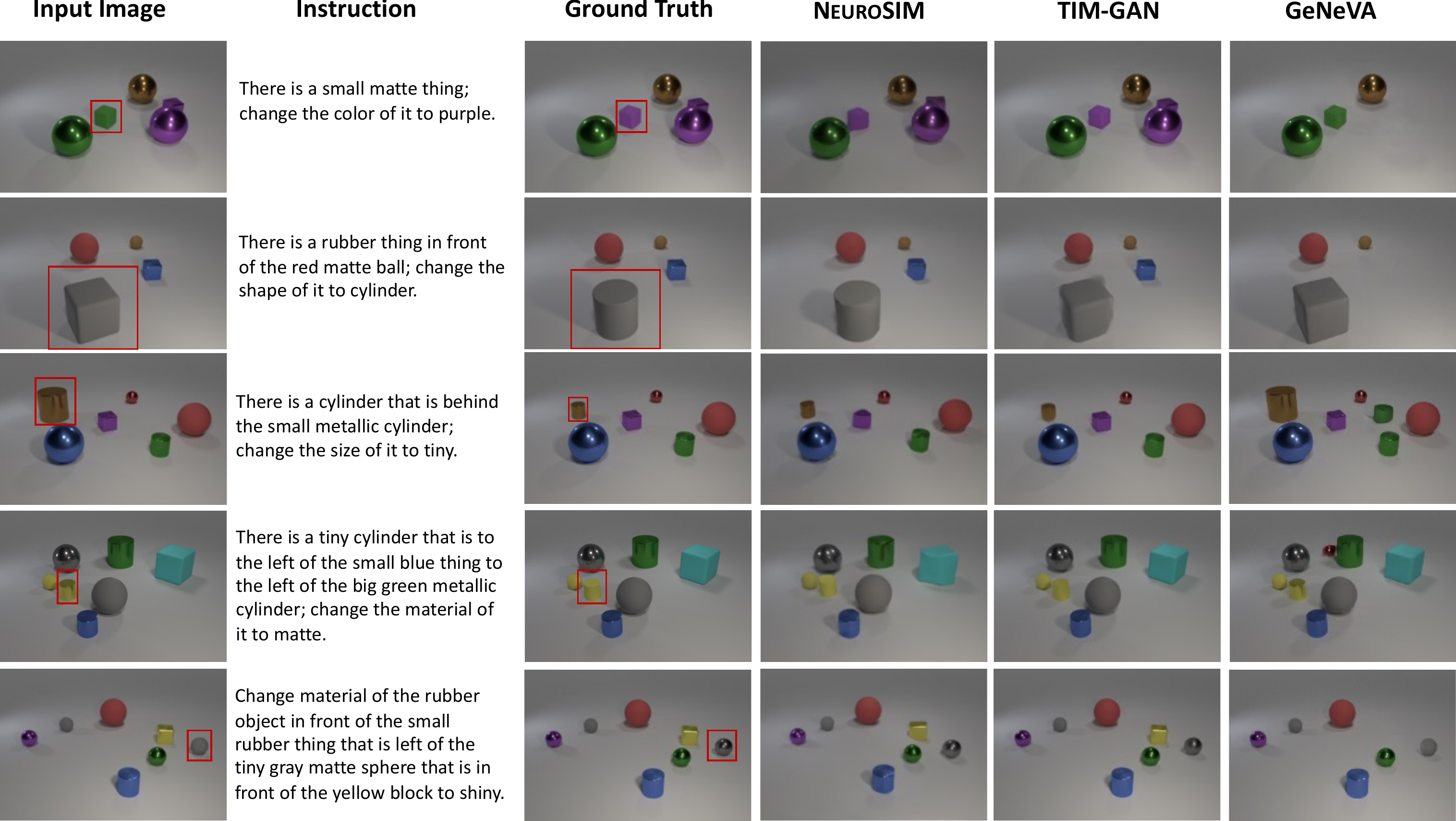}
	\caption{Visual comparison of \method{} with  TIM-GAN and GeNeVA for the \textit{change} operator. The red bounding boxes in the input and ground truth output image indicate the objects required to be changed.}
	\label{fig:app_visual_comparisons_change}
\end{figure*}
\begin{figure*}[h!]
	\centering
	\includegraphics[page=3,width=\textwidth]{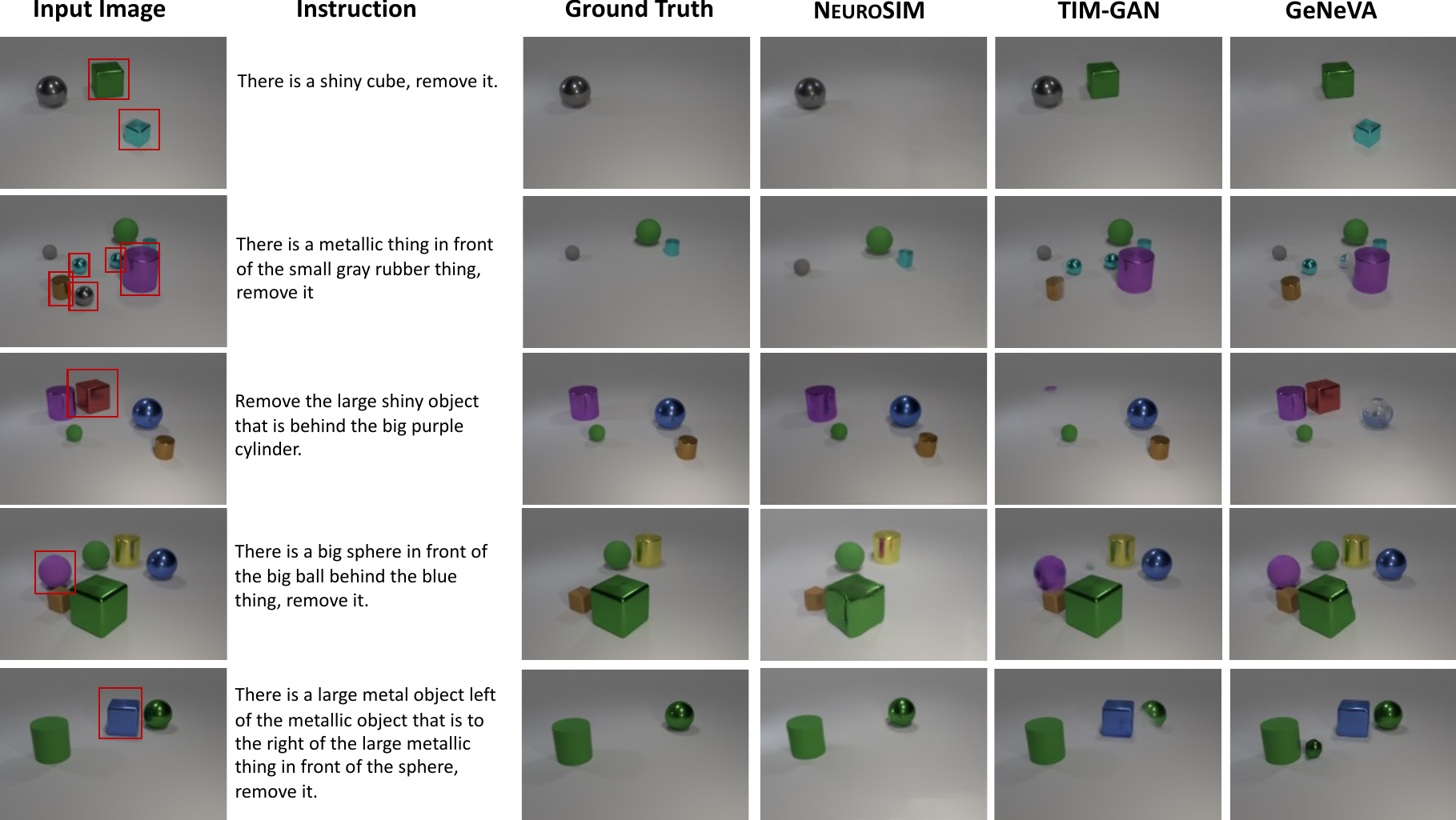}
	\caption{Visual comparison of \method{} with TIM-GAN and GeNeVA for the \textit{remove} operator. The red bounding boxes in the input image indicate objects required to be removed.}
	\label{fig:app_visual_comparisons_remove}
\end{figure*}
Figures \ref{fig:app_visual_comparisons_add}, \ref{fig:app_visual_comparisons_change}, \ref{fig:app_visual_comparisons_remove} compare the images generated by \method{}, TIM-GAN, and GeNeVA on add, change and remove instructions respectively. \method{}'s advantage lies in semantic correctness of manipulated images. For example, see Figure \ref{fig:app_visual_comparisons_add} row \#3,4; Figure \ref{fig:app_visual_comparisons_change} row \#2; \ref{fig:app_visual_comparisons_remove} all images. In these images, \method{} was able to achieve semantically correct changes, while TIM-GAN, GeNeVA faced problems like blurry, smudged objects while adding them to the scene, removing incorrect objects from the scene, or not changing/partially changing the object to be changed. Images generated by TIM-GAN are better in quality as compared to \method{}. We believe the reason for this is that TIM-GAN, being fully supervised, only changes a small portion of the image and has learned to copy a significant portion of the input image directly to the output. However, this doesn't ensure the semantic correctness of TIM-GAN's manipulation, as described above with examples where it makes errors. The images generated by \method{} look slightly worse since the entire image is generated from object based embeddings in the scene graph. Improving neural image rendering from scene graphs can be a promising step to improve \method{}. 
\section{Errors}
\label{app:errors}
\begin{figure*}[h!]
\vspace{-0.5em}
	\centering
	\includegraphics[keepaspectratio,width=\textwidth]{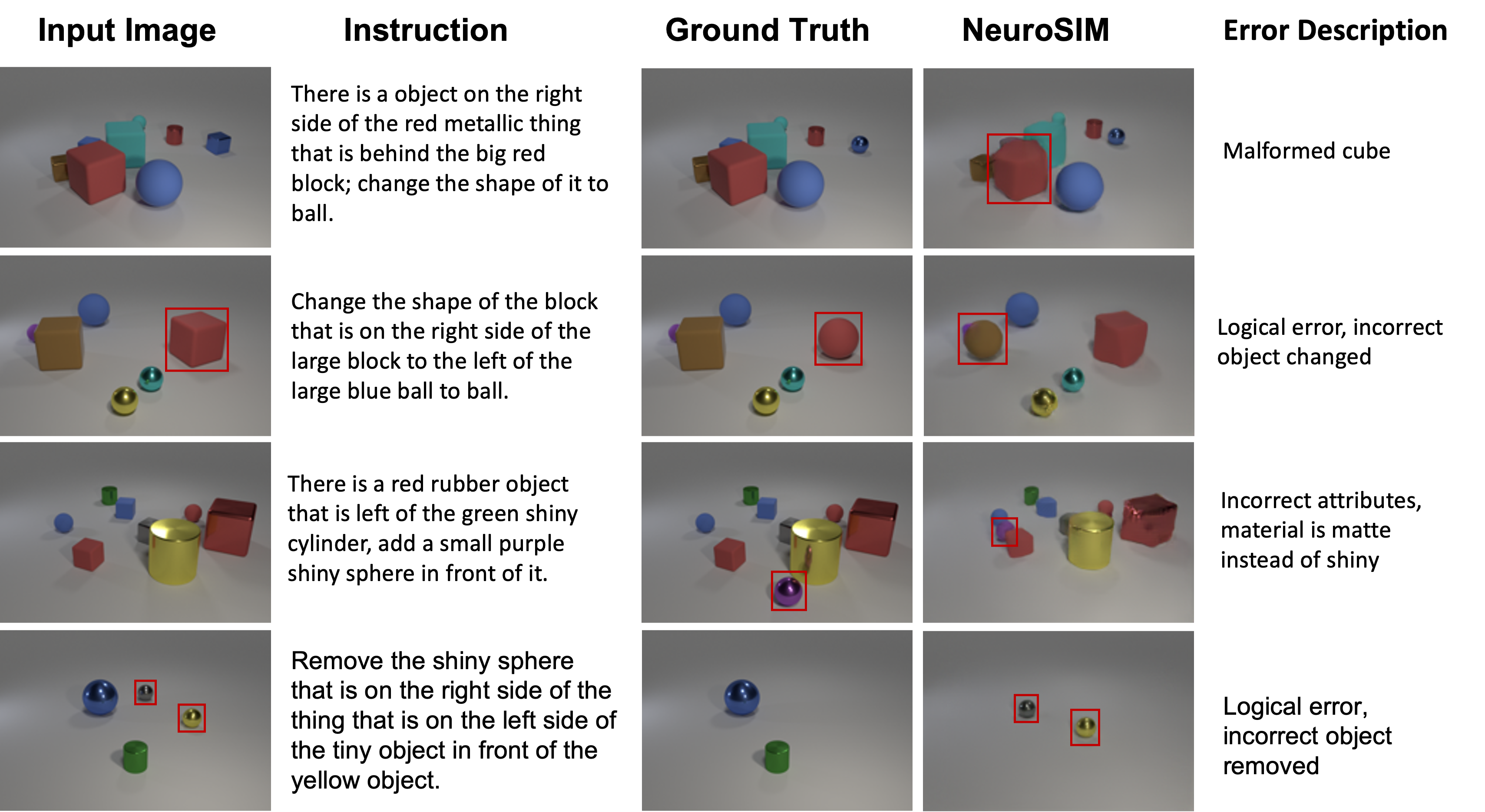}
	\caption{Types of errors in \method{}.}
	\label{fig:visual_comparisons_errors}
	\vspace{-0.5em}
\end{figure*}
Figure \ref{fig:visual_comparisons_errors} captures the images generated by our model where it has made errors. The kind of errors that \method{} makes can be broadly classified into three categories.
\begin{itemize}[leftmargin=*]
\item {\bf [Rendering Errors]} This set includes images generated by our model which are semantically correct but suffer from rendering errors. The common rendering errors include malformed cubes, partial cubes, change in position of objects, and different lighting.
\item {\bf [Logical Errors]} This set includes images generated by our model which have logical errors. That is, manipulation instruction has been interpreted incorrectly and a different manipulation has been performed. This happens mainly due to an incorrect parse of the input instruction into the program, or manipulation network not trained to perfection. For example, {\em change} network changing attributes which were supposed to remain unchanged.
\item {\bf [VQA Errors]} The query networks are not ideal and have errors after they are trained on the VQA task. This in turn causes errors in supervision (obtained from query networks) while training the manipulation networks and leads to a less than optimally trained manipulation network. Also, during inference, object embeddings may not be perfect due to the imperfections in the visual representation network and that leads to incorrect rendering.
\end{itemize}

\section{Ablations}
\label{supp:ablations}
Table \ref{tab:loss_ablations} shows the performance of \method{} when certain loss terms are removed while learning of the networks. This depicts the importance of loss terms that we have considered. In particular we test the performance of the network by removing edge adversarial loss used by add network (row 2), object adversarial losses for both add and change networks (row 3, 5),  self supervision losses used by add network (row 4), cyclic (row 6) and consistency (row 7) losses used by change network.

\begin{table}[ht]
    \setlength\tabcolsep{4pt}
    \renewcommand{\arraystretch}{1.3}
    \setlength\extrarowheight{2pt}
    \centering
    \begin{tabular}{lrr}
        \toprule[0.7pt]
        Loss & R1 & R3\\
        \toprule
        $\ell$ & 45.3 & 65.5\\
        $\ell - \ell^{add}_{\text{edgeGAN}}$ & 43.7 & 66.0\\
        $\ell - \ell^{add}_{\text{objGAN}}$ & 44.3 & 60.2\\
        $\ell - \ell^{add}_{\text{objSup}} - \ell^{add}_{\text{edgeSup}}$ & 44.1 & 57.9\\
        $\ell - \ell^{change}_{\text{objGAN}}$ & 44.9 & 61.5\\
        $\ell - \ell^{change}_{\text{cycle}}$ & 36.5 & 51.1\\
        $\ell - \ell^{change}_{\text{consistency}}$ & 31.0 & 44.8\\
        \bottomrule[1.0pt] 

    \end{tabular}
\caption{Ablations conducted by removing some loss terms. $\ell$ is the total loss before any ablation. For each loss term being removed, the superscript denotes which network it belongs to (add or change). Ablations are conducted for the setting where $\beta = 0.054$ (see main paper Section 4 for the definition of $\beta$)}
\label{tab:loss_ablations}
\end{table}

\section{Interpretability of \method{}}
\label{app:interpretable_programs}
\begin{figure*}[h!]
\vspace{-0.5em}
	\centering
	\includegraphics[keepaspectratio,width=\textwidth]{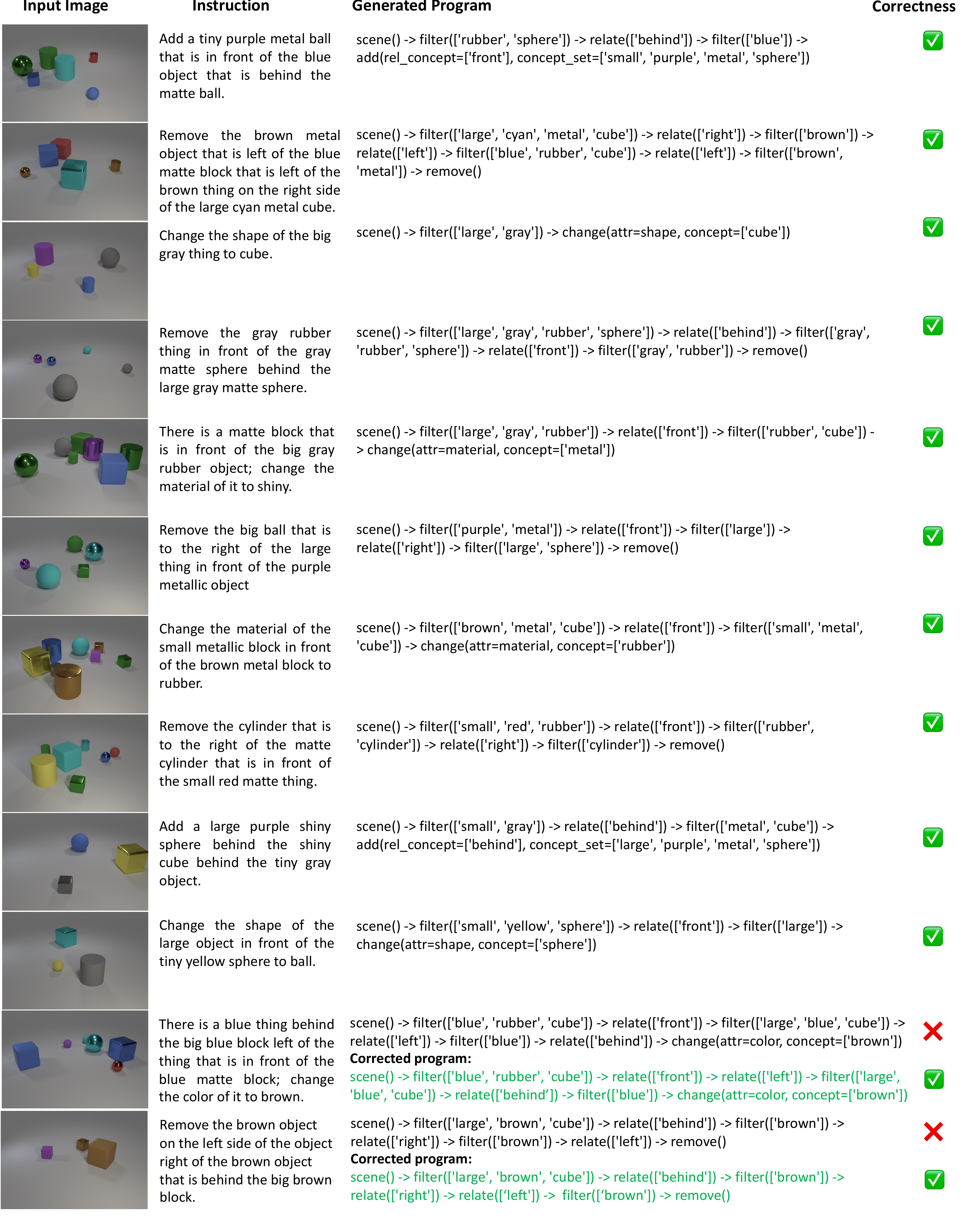}
	\caption{Qualitative examples of generated programs by \method{}.}
	\label{fig:qualitative_gen_programs}
	\vspace{-0.5em}
\end{figure*}

\method{} allows for interpretable image manipulation through programs which are generated as an intermediate representation of the input instruction. This is one of the major strengths of \method{}, since it allows humans to detect where \method{} failed. This is not possible with purely neural models, that behave as a black box. Knowing about the failure cases of \method{} also means that it can be selectively trained to improve certain parts of the network (for eg individually training on change instructions to improve the change command, if the model is performing poorly on change instructions).
We now assess the correctness of intermediate programs using randomly selected qualitative examples present in Figure \ref{fig:qualitative_gen_programs}. Since no wrong program was obtained in the randomly selected set, we find 2 more data points manually, to show some wrong examples.

\section{Human Evaluation Details}
\label{hum_eval_appendix_with_details}

\begin{figure*}[ht!]
	\centering
	\subfloat{%
		\includegraphics[page=5,width=0.64\textwidth]{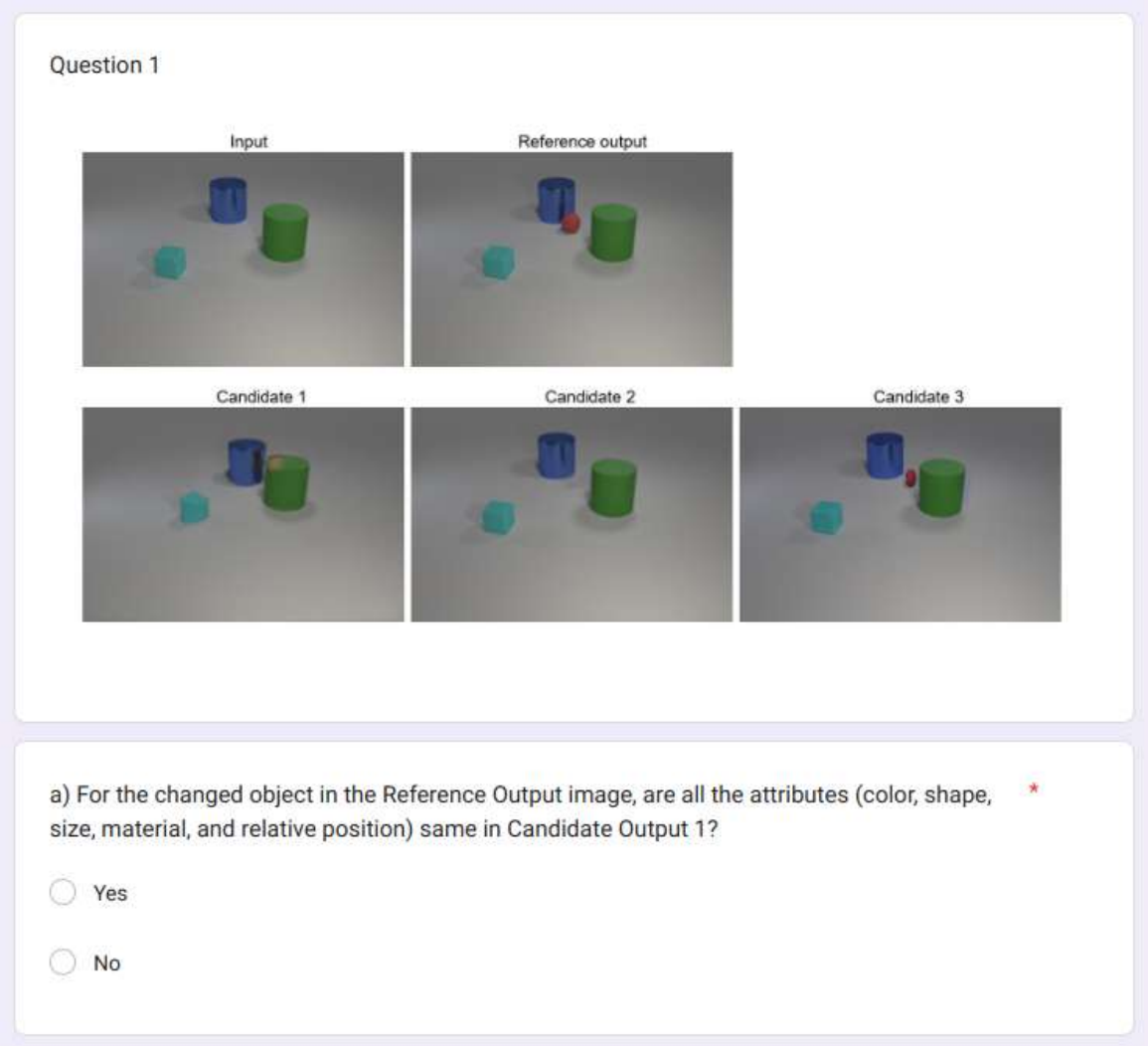}%
	}\par
	\centering
	\subfloat{%
		\includegraphics[page=6,width=0.64\textwidth]{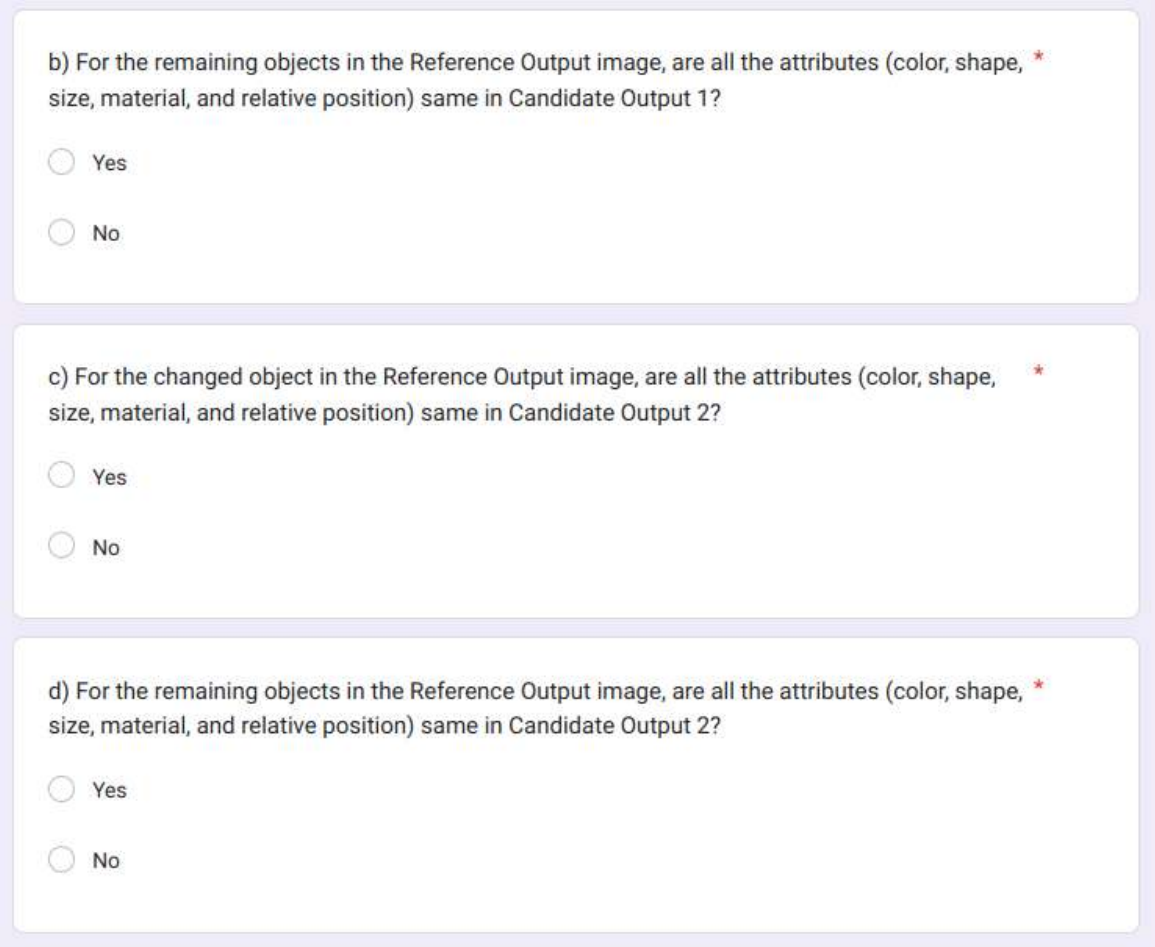}%
	}\par
        \centering
	\subfloat{%
		\includegraphics[page=7,width=0.64\textwidth]{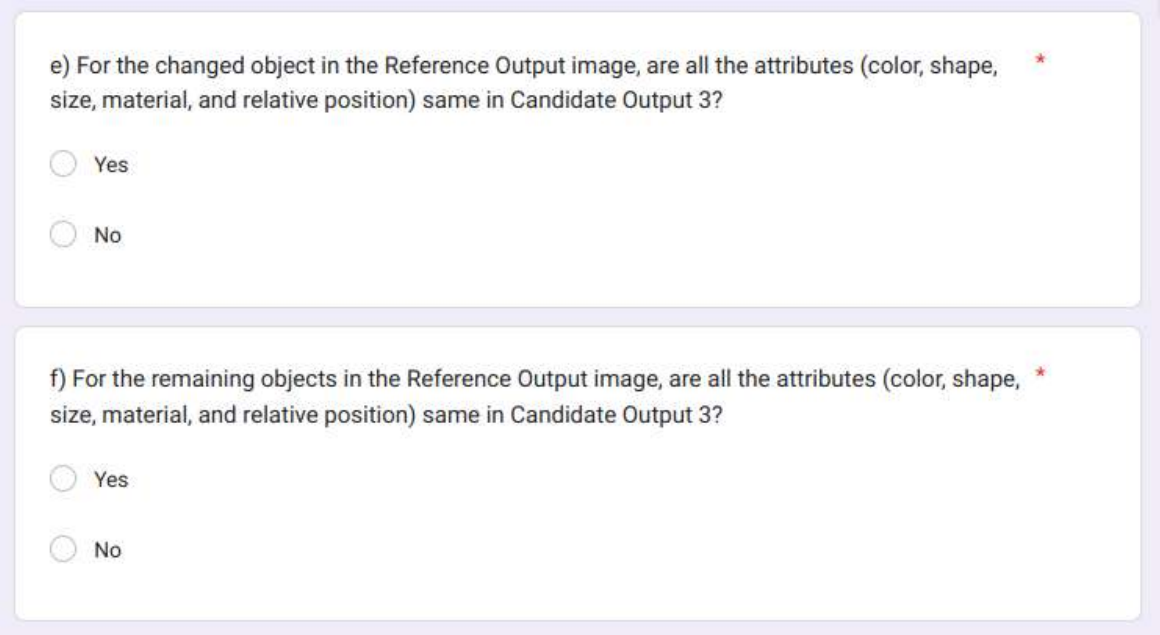}%
	}\par
	\caption{Screenshots for human evaluation study. See Section \ref{app:hum_eval} for more details}
	\label{fig:hum_eval_screenshots}
\end{figure*}

See Table~\ref{human_eval_questions} for the questions (paraphrased) asked to the evaluators.  Detailed instructions and an example of the questions provided to the evaluators can be found in Figure \ref{fig:hum_eval_screenshots}. A total of $10$ evaluators, consisting of a mix of undergraduate and post-graduate students, were involved in the study. The same set of $30$ random images were given to each evaluator. They were compensated at a rate three times the average hourly salary in the country of origin. Each evaluator was given upto $24$ hours to complete the task.

\section{Simplifying Multi-Hop Instructions using NeuroSIM Modules}
\label{simplifiy_MH}
In this section, we provide details on our method of utilizing the trained semantic parser to convert the complex multi-hop instruction into a simplified 0 or 1 hop instruction. We generate three simplified templates one for each edit operation.\\
1. \textit{Change the [attribute] of [size] [color] [material] [shape] to [attribute']}\\
2. \textit{Remove the [size] [color] [material] [shape]}\\
3. \textit{Add a [size] [color] [material] [shape] to the [relation] of [shape']}. Next, given a multi-hop instruction we parse it using our semantic parser which gives us the object's embedding on which either an operation is to be executed (in case of change and remove operations) or a new object has to be inserted in relation to it (in case of add operation). The trained query-networks predicts the symbolic values of the concepts in the placeholders. Example, if the MH instruction is "\textit{Change the size of the big thing that is behind the metallic cylinder behind the purple object that is to the right of the big brown shiny object to tiny}" , we find the placeholder attributes to be operation=change, attribute=size, color=yellow, shape=cube, size=large, material=rubber, attribute'=tiny. Hence the simplified instruction becomes, "\textit{Change the size of the large yellow rubber cube to tiny}". Add and Remove instructions follow similarly.

\begin{table*}[ht]
\centering
\setlength\tabcolsep{4pt} 
\renewcommand{\arraystretch}{1.3}
\setlength\extrarowheight{2pt}
{
\begin{tabular}{p{2.5cm}p{11cm}}
\toprule

\multirow{7}{*}{{\texttt{Question 1:}}} &  \textbf{[Change]} Are all the attributes (color, shape, size, material, and relative position) of the changed object mentioned in the instructions identical between the ground truth image and the system-generated image? \\

&  \textbf{[Add]} Are all the attributes (color, shape, size, material, and relative position) of the added object mentioned in the instructions identical between the ground truth image and the system-generated image? \\

& \textbf{[Remove]} Are same objects removed in ground truth image and the system-generated image?  \\

\midrule
\multirow{7}{*}{{\texttt{Question 2:}}} & \textbf{[Change]} Are all the attributes (color, shape, size, material, and relative position) of the remaining objects identical between the ground truth image and the system-generated image? \\

& \textbf{[Add]} Are all the attributes (color, shape, size, material, and relative position) of the remaining objects identical between the ground truth image and the system-generated image? \\

& \textbf{[Remove]} Are all the attributes (color, shape, size, material, and relative position) of the remaining objects identical between the ground truth image and the system-generated image?  \\

\bottomrule

\end{tabular}
}
\caption{Questions asked to human evaluators for evaluating \method{} and TIM-GAN. Note that there are some variations in the questions for Change, Add, and Remove instructions dues to different semantic nature of the instructions.}
\label{human_eval_questions}
\end{table*}

\begin{figure*}[ht!]
	\centering
	\subfloat{%
		\includegraphics[page=1,width=0.7\textwidth]{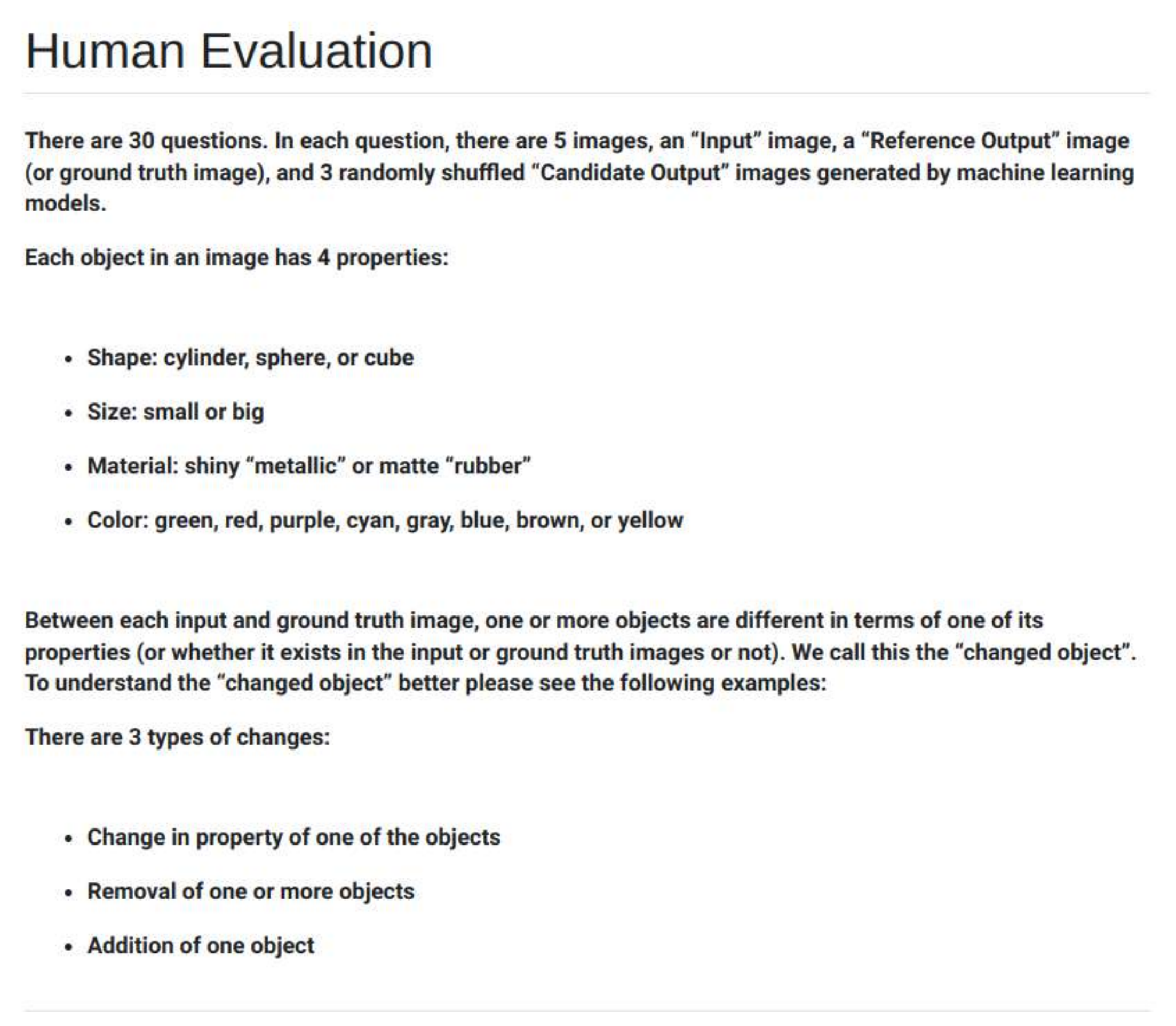}%
	}\par
	\centering
	\subfloat{%
		\includegraphics[page=2,width=0.7\textwidth]{./figures/human_eval.pdf}%
	}\par
	\label{fig:hum_eval_screenshots0}
\end{figure*}
\begin{figure*}[ht!]
	\centering
	\subfloat{%
		\includegraphics[page=3,width=0.7\textwidth]{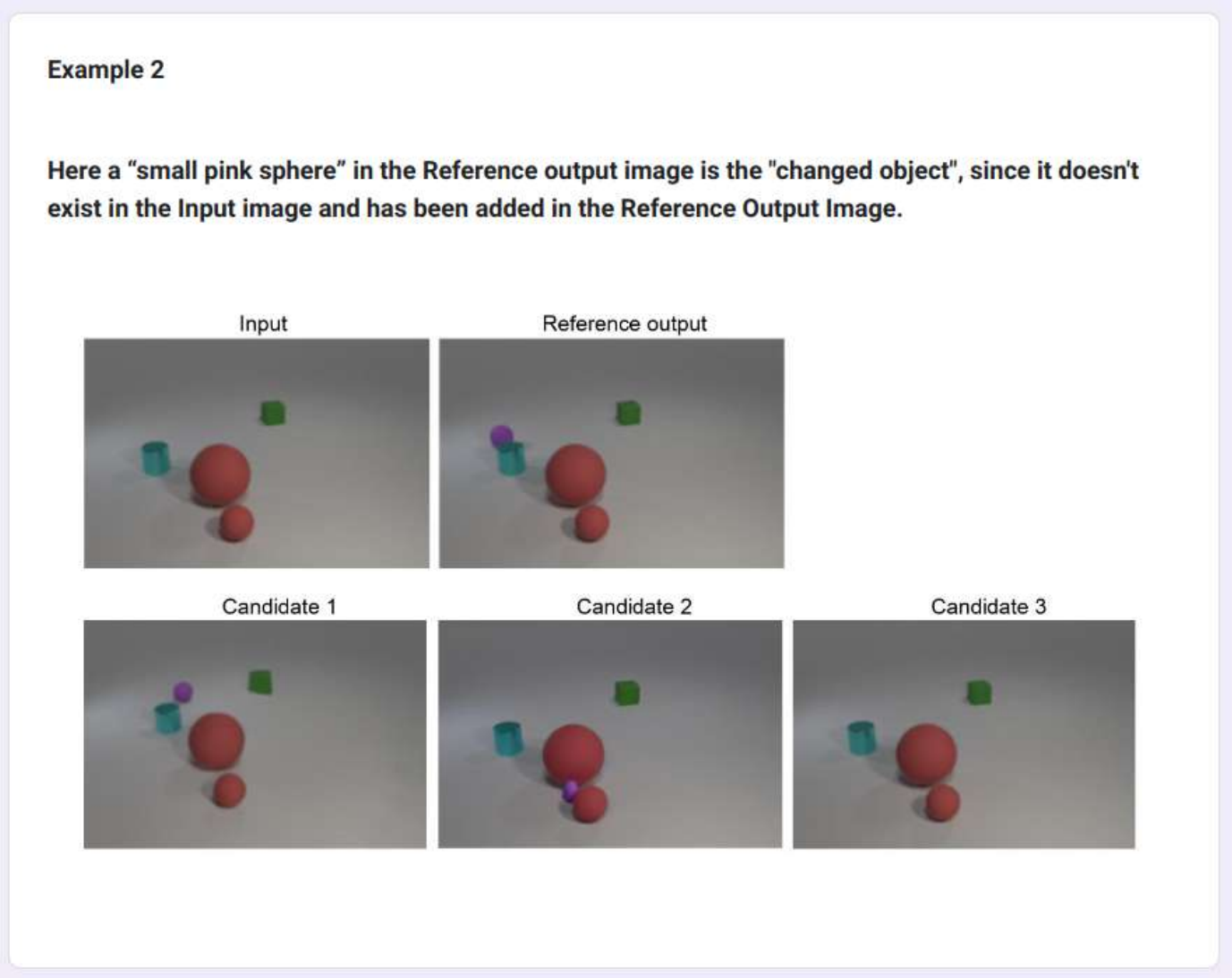}%
	}\par
	\centering
	\subfloat{%
		\includegraphics[page=4,width=0.7\textwidth]{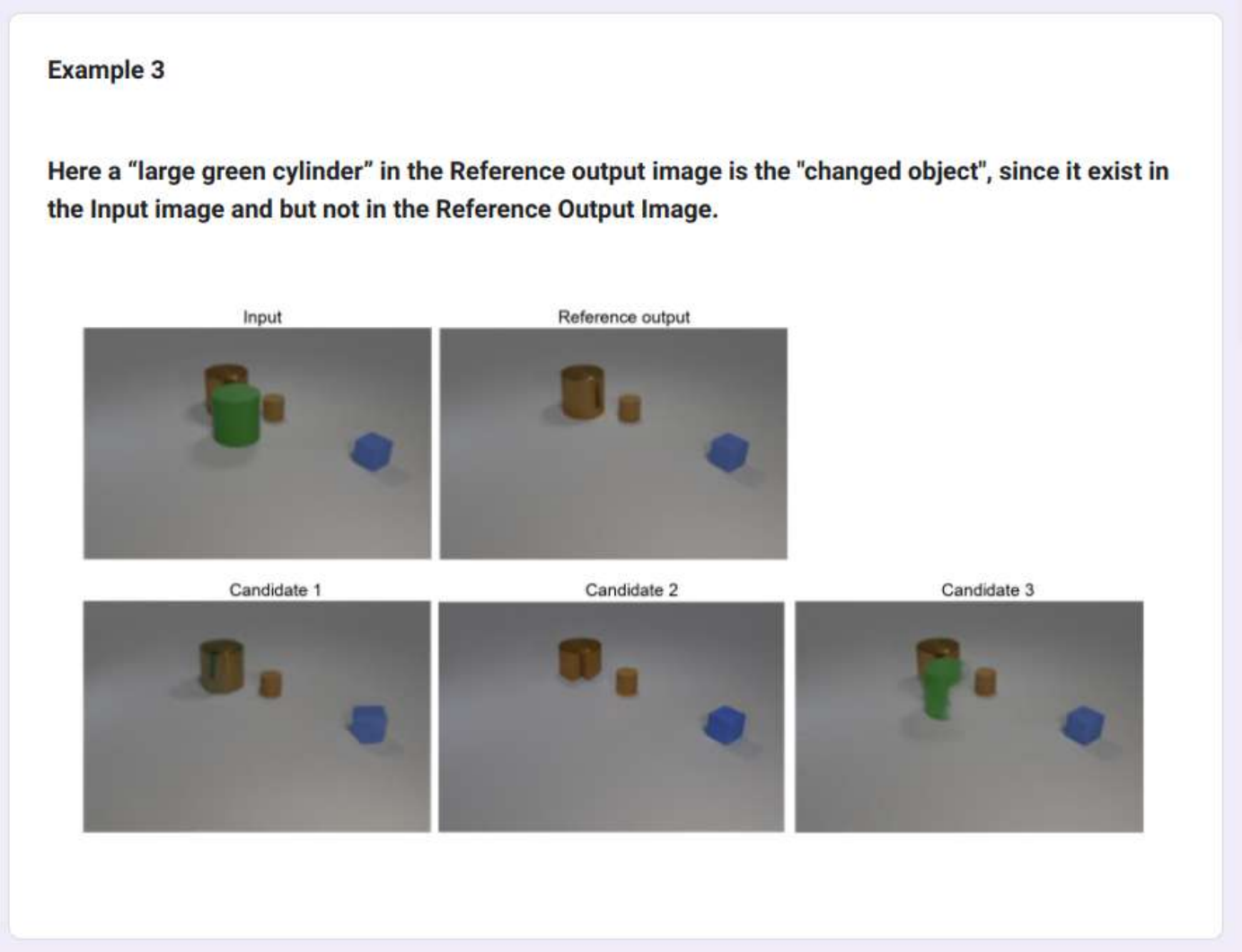}%
	}\par
\end{figure*}
